\documentclass[a4paper,10pt]{scrartcl}

\usepackage{amsmath}
\usepackage{amssymb}
\usepackage{subfig}
\usepackage{graphicx}

\usepackage{tikz}

\usepackage{tabularx}

\usepackage{booktabs}
\usepackage{dcolumn}
\usepackage{multirow}
\usepackage{siunitx}
\usepackage[round]{natbib}

\usepackage[margin=2.5cm]{geometry}

		            \usepackage{url}
\usepackage{bm}                         

\newcommand{\fref}[1]{\protect\subref{#1}}

\renewcommand{\vec}{\boldsymbol}
\renewcommand{\L}{\mathcal{L}}
\newcommand{\D}{\mathcal{D}}
\newcommand{\Exp}{\mathbb{E}}
\newcommand{\rX}{\bm{X}}                \newcommand{\rY}{\bm{Y}}                
\newcommand{\x}{\vec{x}}
\newcommand{\y}{\vec{y}}
\newcommand{\ypred}{\vec{\hat{y}}}
\newcommand{\ymode}{\vec{y^{\star}}}
\newcommand{\yp}{\hat{y}}

\newcommand{\p}{\boldsymbol P(\rY|\x)}
\renewcommand{\P}{\boldsymbol P}

\newcommand{\Sec}[1]{Section \ref{#1}}
\newcommand{\Eq}[1]{equation~{\eqref{#1}}}

\newcommand{\Fig}[1]{Figure~\ref{#1}}
\newcommand{\Tab}[1]{Table~\ref{#1}}

\newcommand{\sB}{\{0,1\}}       \newcommand{\dataset}{}

\DeclareMathOperator*{\argmax}{arg\,max}

\newcolumntype{Y}{>{\raggedleft\arraybackslash}X}
\newcolumntype{Z}{>{\centering\arraybackslash}X}
\newcommand{\zspace}{\quad}

\newcounter{AnalysisNumber}

\begin{document}

\title{Estimating Multi-label Accuracy using Labelset Distributions}

\author{Laurence A.~F.~Park\thanks{Laurence Park is with the
    School of Computer, Data and Mathematical
    Sciences, Western Sydney University, Australia
   \texttt{lapark@westernsydney.edu.au}}
  \and Jesse Read\thanks{Jesse Read is with LIX, Ecole Polytechnique, Institut
    Polytechnique de Paris, France
   \texttt{jesse.read@polytechnique.edu}}}

\date{}

\maketitle

\begin{abstract}

A multi-label classifier estimates the binary label state (relevant vs irrelevant) for each of a set of concept labels, for any given instance. 
Probabilistic multi-label classifiers provide a predictive posterior distribution over all possible labelset combinations of such label states (the powerset of labels) from which we can provide the best estimate, simply by selecting the labelset corresponding to the largest expected accuracy, over that distribution. For example, in maximizing exact match accuracy, we provide the mode of the distribution. 
But how does this relate to the confidence we may have in such an estimate? Confidence is an important element of real-world applications of multi-label classifiers (as in machine learning in general) and is an important ingredient in explainability and interpretability. However, it is not obvious how to provide confidence in the multi-label context and relating to a particular accuracy metric, and nor is it clear how to provide a confidence which correlates well with the expected accuracy, which would be most valuable in real-world decision making. 
In this article we estimate the expected accuracy as a surrogate for confidence, for a given accuracy metric. We hypothesise that the expected accuracy can be estimated from the multi-label predictive distribution. 
We examine seven candidate functions for their ability to estimate expected accuracy from the predictive distribution. We found three of these to correlate to expected accuracy and are robust. Further, we determined that each candidate function can be used separately to estimate Hamming similarity, but a combination of the candidates was best for expected Jaccard index and exact match.
\end{abstract}

\newcommand{\downthree}{$\downarrow\downarrow\downarrow$}
\newcommand{\downtwo}{$\downarrow\downarrow$}
\newcommand{\downone}{$\downarrow$}
\newcommand{\upthree}{$\uparrow\uparrow\uparrow$}
\newcommand{\uptwo}{$\uparrow\uparrow$}
\newcommand{\upone}{$\uparrow$}

\section{Introduction}
\label{sec:intro}

Multi-label classifiers model the relationship between an observation vector and a set of label concepts. 
Multi-label classification is relevant to numerous domains, including
categorising documents, tagging images, assigning medical diagnoses,
and bioinformatics.  A review of some well-known multi-label methods
is given in \citet{Review}, and more generally a unifying view (with the general case of multi-output learning) is provided by \citet{UnifiedView}.

Formally, given a set of $L$ label concepts
$\L = \{1,2,\ldots,L\}$, a labelset can be denoted $\y = [y_1,\ldots,y_L] \in \{0,1\}^L$, where $y_j = 1$ implies that the $j$-th label is relevant to the corresponding input vector $\x$ (and $y_j=0$ implies that it is not relevant). In other words, $\y$ equivalently denotes a subset of label concepts. There are $2^L$ such possible labelsets (i.e., combinations of labels); i.e., the size of the powerset of $\L$ and the domain $\{0,1\}^L$.

For any input vector $\vec{x}$, a probabilistic multi-label classifier provides a distribution $\P(\rY|\x)$ over all $2^L$ elements of the powerset. This provides a useful measure of confidence $P(\ypred|\x)$ for any given prediction $\ypred$, so that a practitioner may have some insight into the reliability of the prediction. 

Our notation is provided in \Tab{tab:notation}. We further clarify with two toy examples as follows, which related to possible applications of multi-label classifiers. 

\begin{table}[h]
	\centering
	\caption{Notation and terminology used throughout the article}
        \label{tab:notation}
        \begin{tabular}{|lp{25em}|}
			\hline
			Symbol & Meaning \\
			\hline
				$\mathcal{L}$ &  \emph{set} of \emph{label} concepts, $\L=\{1,2,\ldots,L\}$          \\
				$y_j$ & label relevance, $y_j \in \{0,1\}$; where $y_j = 1 \Leftrightarrow $ the $j$-th label is relevant \\
				$\y$ &  \emph{labelset}, $\y \in \sB^L$, e.g., $\y = [0,1,0,1]$ if $2$nd and $4$th labels are relevant \\
				$\y_i$ &  $i$-th labelset, $i=1,\ldots,2^L$, $\y_i \in \{0,1\}^L$, $\y_i = [y_{i1},\ldots,y_{iL}]$ where $y_j \in \{0,1\}$ \\
				$\x^{(n)}$ &  $n$-th \emph{instance} (of $N$ in total) from the dataset \\
				$\y^{(n)}$ &  labelset associated with $n$-th instance ($N$ in total), $\y^{(n)} = [y^{(n)}_1, \ldots,y^{(n)}_L]$         \\
				$\ypred$ &  any prediction from a multi-label classifier, $\ypred \in \{\y_i\}_{i=1}^{2^L}$ \\
				$P(\y_i|\x)$ & shorthand for $P(\rY=\y_i|\rX=\x)$; if conditional is clearly implied, then $P(\y_i)$ \\
				$\p$ & shorthand for $P(\rY|\rX=\x)$; the full distribution over all labelsets ($\{P(\y_i|\x) \mid \forall i\}$)   \\
				$\ymode$ &  MAP/mode estimate; see \Eq{eq:MAP} \\
$\mathcal{S}(\y_i,\y)$ & a \emph{similarity function}/accuracy metric; see, e.g., equations \eqref{eq:exact}, \eqref{eq:hamming}, and \eqref{eq:jaccard} \\
				$E_{\mathcal{S}}(\y_i)$ & the \emph{expected accuracy} of predicting $\y_i$ under metric $\mathcal{S}$; see \Eq{eq:Exp} \\
				$f(\p)$ & a candidate \emph{score function}; confidence $\in [0,1]$ of predicting under $\p$ \\
			\hline
		\end{tabular}
\end{table}

\paragraph{Example 1} Given a patient profile $\x$ the task is to estimate if the patient has suffered from any of a number of ailments $\L = \{\textsf{Diabetes}, \textsf{Hypertension}, \textsf{SARS-CoV-2}\}$. We want to output a prediction $\ypred  = [\yp_1,\yp_2,\yp_3]$ to apply to a patient's medical record, for future reference. For example, if $\yp_3 = 1$ there is a strong belief that the patient has in the past contracted the covid virus. It is important to have a quantification of how strong that belief is, and how it relates to the expected accuracy. How confident should we be that each of the three assessments -- $\yp_1,\yp_2$, and $\yp_3$ -- is correct?

\paragraph{Example 2} Consider the same set of categories as in the previous example, but label-vector $\ypred$ refers to an estimate of the \emph{current} state of the patient which will be used as a tool for diagnosis. In this case, the accurate detection of combinations alongside the patient profile $\x$ can be crucial. The diseases represented by the first two labels are comorbidities for covid, and therefore it is of utmost importance to detect the correct \emph{combination}, especially for certain factors represented in $\x$, e.g., indicating an elderly patient. Therefore, we are concerned about the confidence on the predicted combination, and how it might relate to a metric that evaluates labels is combination.

\medskip

In both examples, beyond a classification $\ypred$, we desire some confidence value $f$ associated with such a classification; such that -- for example -- a value from $f$ near $0$ represents low confidence, and near $1$ a classification which is, according to the model, quite certain and likely to be correct under the relevant accuracy metric. In the first example, that metric would be something like Hamming similarity, and in the second, something like exact match similarity; both discussed below. Having such a gauge of confidence for a chosen metric is increasingly important, and often essential in practical machine learning, allowing a practitioner to gain trust in the predictions provided by the model. Particularly, in many domains, such as legal and medical applications, this is crucial \cite{InterpretableML}. 

Is is clear, and well-known in the multi-label literature, particularly that dealing with probabilistic models, that join distribution $P(\y|\x)$ is readily available from which to derive a confidence bound. However, it is not clear which is the best way to derive that confidence; in particular, with regard to different evaluation metrics. In Example 1, we might use Hamming Similarity. In Example 2, we might use Exact Match. The confidence of our prediction should be different depending on what we are predicting. 

In the literature, typically the labelset $\ymode$ associated to the mode of the multi-label distribution,
\begin{equation}
	  \label{eq:MAP}
	\ymode = \argmax_{\y \in \{0,1\}^L} P(\y|\x)
\end{equation}
is selected as the predicted response (including, implicitly, in non-probabilistic classifiers), and we would report $P(\ymode|\x)$ as the \emph{confidence}. The remainder of the distribution is typically discarded.

Numerous methods from the literature have been proposed can produce all ingredients of \Eq{eq:MAP}; offering some mechanism to deal with the inherent complexity. For example, the class of methods often called the `\emph{label powerset} approach' \citep{RAkEL2,MetaLabels}, which implicitly replaces the powerset $\{0,1\}^L$ with a smaller subset of much smaller dimension. The family of \textit{classifier chains} \citep{CCReview} offers search mechanisms to efficiently traverse parts of the space with a greedy exploitation of the gain rule \cite{ECC2} or some other approximation \cite{PCCInferenceSurvey}. Even when the base classifiers (responsible for individual label predictions) are not inherently probabilistic, such as a decision tree, an approximation to $\p$ can easily be obtained via, e.g., ensemble voting.

Note that in this paper we do not focus on the issue of computational complexity, but rather take an interest in the confidence predictions themselves and more generally, the posterior joint distribution they provide: how confident can we be regarding any given prediction $\ypred$, and specifically the mode $\ymode$? In other words, given the posterior distribution $\p$ provided by our classifier, with what confidence can we predict $\ymode$, with regard to our chosen accuracy metric? What is the \emph{expected accuracy} of our prediction? We answer this question in terms of the \emph{shape} of $\P(\y|\x)$, and we present seven candidate confidence functions to produce an estimate. We confirm with an empirical investigation on a range of multi-label datasets and a number of different methods.

The article will proceed as follows: Section \ref{sec:pml} gives an
introduction to probabilistic multi-label learning and describes the
research problem of confidence in multi-label predictions. In Section
\ref{sec:related} we survey and discuss existing approaches in the
literature and connect them to our contribution. In Section
\ref{sec:acceptance} we introduce the set of candidate
functions that have the potential to measure expected
  accuracy. The potential of these functions to measure the
expected accuracy for multi-label prediction is examined throughout this article. Sections \ref{sec:Experiments}--\ref{sec:absolute} provide
experimental results and discussion, examining the effectiveness of
each candidate expected accuracy function and the effect of data and methods on each.  Final concluding remarks and speculation of promising future work is given in \Sec{sec:conclusion}.

\section{Confidence for Probabilistic Multi-label Learning}
\label{sec:pml}

In this section we provide background for discussing confidence in the context of multi-label learning. Later in \Sec{sec:related} we will address some other relevant prior work, and in \Sec{sec:acceptance} we develop the candidate functions that we analyse specifically in this work. 

\subsection{Confidence in binary classification}

The confidence of a probabilistic binary (single-label) classifier in its prediction $\yp$ for some test instance $\x$, is the expected accuracy for providing that prediction versus the unknown true label $y$. If this accuracy metric is classification accuracy,
\[
	\mathcal{S}(\yp,y) = \left[\yp=y\right]
\]
where $\left[ A \right]$ is an indicator function, returning $1$ if $A$ is true and $0$ otherwise; i.e., $\mathcal{S}(\yp,y)$ provides $1$ iff the prediction $\yp$ is identical to the true label $y$. Then our confidence is 
\begin{equation}
	E_\mathcal{S}(\yp) = \Exp[\mathcal{S}|\x] = \mathbb{E}_{Y\sim P}[\left[\yp=Y\right]|\x] = \left(\sum_{y \in \{0,1\}} 1_{y=\yp} \cdot P(Y =y |\x) \right) = P(Y=\yp|\rX = \x) 
	\label{eq:binary_conf}
\end{equation}
i.e., the expected accuracy; expected value of $\mathcal{S}$. Because we do not (at testing time, in practice) know the true value of $y$, we only have the conditional distribution $P(Y|\x)$ is provided by the probabilistic binary classifier. 

For example, if 
\[
	P(Y=\yp|\rX=\x) = 0.6
\]
we can understand that the model is 60\% confidence; or has a 60\% chance of the making a correct 
prediction $\yp$ for that given observation $\x$; and thus that the expected accuracy is $0.6$. 

High confidence is equivalent to low uncertainty. The uncertainty is expressed in the form of the expectation, which arises because we have no knowledge of the true $y$ that random variable $Y$ takes here. 

\subsection{Confidence in multi-label classification}

A probabilistic multi-label classifier provides
\[\P(\rY|\rX = \x)\]
and thus $P(\rY = \y_i|\rX = \x)$ for any $\y_i \in \{0,1\}^L$, i.e., the probability of each label state $\y_i$ to be applied to instance $\x$. 
We provide an example in \Tab{tab:example} (recall, also, the summary of notation in Table~\ref{tab:notation}).
An excellent introduction and analysis in the multi-label context (for readers not already familiar) is provided by \citet{OnLabelDependence2}. 

\begin{table}[th]
	\caption{Example of a distribution $\P(\y|\x)$ under an observation $\x$ where $L=3$, i.e., $\{P(\y_i | \x) | i = 1,\ldots,2^L\}$, showing each $\y_i = [y_1,y_2,y_3]$. This distribution is also shown graphically in Figure~\ref{fig:dist-example.e}. Marginal probabilities for each label, $P(Y_j|\x)$, are shown in the final row. The expected accuracy $\Exp[\mathcal{S}]$ for the three different metrics $\mathcal{S}$ we study, are also appended alongside each labelset/prediction. }
  \label{tab:example}
  \label{tab:example2}
  \centering
  \begin{tabular}{r cccc ccc}
    \toprule
	  $i$         & $[y_1$ & $y_2$  & $y_3]$ & $P(\rY = \y_i|\rX = \x)$ & $\mathbb{E}[\text{HS}]$ & $\mathbb{E}[\text{EM}]$ & $\mathbb{E}[\text{JS}]$    \\
    \midrule                                                                                                                              
    1             & $0$    & $0$    & $0$    & 0.000  & \textbf{0.611}          & 0.000                   & 0.000                      \\    
    2             & $0$    & $0$    & $1$    & 0.333  & 0.500                   & \textbf{0.333}          & 0.333                      \\    
    3             & $0$    & $1$    & $0$    & 0.250  & 0.556                   & 0.250                   & 0.333                      \\    
    4             & $0$    & $1$    & $1$    & 0.000  & 0.444                   & 0.000                   & 0.347                      \\    
    5             & $1$    & $0$    & $0$    & 0.250  & 0.556                   & 0.250                   & 0.333                      \\    
    6             & $1$    & $0$    & $1$    & 0.000  & 0.444                   & 0.000                   & 0.347                      \\    
    7             & $1$    & $1$    & $0$    & 0.167  & 0.500                   & 0.167                   & \textbf{0.417}             \\    
    8             & $1$    & $1$    & $1$    & 0.000  & 0.389                   & 0.000                   & 0.389                      \\    
    \midrule
	  $P(y_j|\x)$ & $0.42$ & $0.42$ & $0.33$ &  & & & \\
    \bottomrule
  \end{tabular}
\end{table}

In multi-label classification, there is the complication of many accuracy metrics being used in the community\footnote{In this article we use the term `accuracy' in the generic sense of comparing two labelset predictions under similarity function $\mathcal{S}(\cdot,\cdot)$; rather than a specific sense of some multi-label papers where `accuracy' is synonymous with a particular metric, often Jaccard Similarity}. To compute confidence for multi-label classification, we must compute the expected accuracy with respect to the chosen metric.

One may immediately verify that examining the marginal and joint modes shown in \Tab{tab:example} refer to different tasks/objectives, since the marginal modes indicate a suitable prediction of $\ypred = [0,0,0]$ (with a threshold of $0.5$), whereas the joint mode suggests $\ymode=[0,0,1]$. Further discussion is given in, e.g., \cite{CCAnalysis}. 

Here are the similarity metrics $\mathcal{S}$ that we consider in this paper, based on their
importance and popularity in the multi-label literature (they can be
found throughout many papers, e.g., \cite{ECC2,RAkEL2,park2018.1} and references
therein), defined over a set of $N$ instances and $L$ labels:

\label{sec:sim}
\begin{equation}
	\label{eq:hamming}
\textsc{Hamming Similarity (HS)} := \frac{1}{NL}\sum_{n=1}^{N}\sum_{j=1}^{L}{ \big[ y_{j}^{(n)} = \hat{y}_{j}^{(n)} \big] }\text{,}
\end{equation}
\begin{equation}
	\label{eq:exact}
\textsc{Exact Match (EM)} := \frac{1}{N}\sum_{n=1}^{N}{ \big[ \y^{(n)} = \ypred^{(n)} \big]}\text{, and }
\end{equation}
\begin{equation}
	\label{eq:jaccard}
	\textsc{Jaccard Similarity (JS)} := \frac{1}{N}\sum_{i=1}^{N} \frac{| \vec{y}^{(n)} \wedge \vec{\hat{y}}^{(n)} |}{| \vec{y}^{(n)} \vee \vec{\hat{y}}^{(n)}|}.
\end{equation}
where $\y^{(n)}$ is the labelset associated to the $n$th instance $\x^{(n)}$, 
where $\wedge$ and $\vee$ are the bitwise logical \textsc{and} and \textsc{or} operations, respectively. Other notation is clarified in Table~\ref{tab:notation}.

The expected multi-label accuracy, extending Eq.~\eqref{eq:binary_conf}, is
\begin{equation}
	E_{\mathcal{S}}(\y_i) = \Exp[\mathcal{S}|\x] = \mathbb{E}_{\rY\sim\p}[\mathcal{S}(\y_i, \rY)|\x] = \left( \sum_{\y_j \in \{0,1\}^L} P(\y_j|\x) \mathcal{S}(\y_i, \y_j) \right)
	\label{eq:Exp}
\end{equation}
which produces the final columns in Table~\ref{tab:example2}, according to each of the metrics of our study.

And so, the multi-label classification task may be posed as 
\begin{equation}
  \label{eq:ysim}
	\ypred = \argmax_{\y_i \in \{0,1\}^L} E_\mathcal{S}(\y_i) \end{equation}

\label{sec:models}

As clearly visible in Table~\ref{tab:example2}, the labelset with the greatest similarity [to the true labelset] 
(shown in bold) depends on that similarity/accuracy function.

Note that by using Exact Match, we obtain the mode of the
labelset distribution,
\begin{align*}
  \ypred &= \argmax_{\y_i \in \{0,1\}^L} \mathbb{E}_{\rY\sim\p}[\text{EM}(\y_i, \rY) | \x] \\
         &= \argmax_{\y_i \in \{0,1\}^L} \left ({\sum_{\y_j \in \{0,1\}^L} \text{EM}(\y_i, \y_j)P(\y_j|\x)}\right ) \\
         &= \argmax_{\y_i \in \{0,1\}^L} P(\y_i|\x) = \ymode
\end{align*}
therefore, equation \ref{eq:ysim} is equivalent to equation
\ref{eq:MAP} when using Exact Match. This is also seen in Table~\ref{tab:example}.

If we were able to assume independent labels we can expand $P(\y_i|\x)$ from  \Eq{eq:MAP} as a product of independent elements and move the $\argmax$ inside, to obtain
\begin{align}
	\ypred &= \argmax_{y_1,\ldots,y_L} \big[ P(y_1|\x) \cdot \cdots \cdot P(y_L|\x) \big] \notag \\
		  &= \big[ \argmax_{y_1 \in \{0,1\}} P(y_1|\x),\ldots,\argmax_{y_L \in \{0,1\}} P(y_L|\x) \big] \label{eq:tag}
\end{align}

If using Hamming Similarity, we obtain,
\begin{align*}
  \ypred &= \argmax_{\y_i \in \{0,1\}^L} \mathbb{E}_{\rY\sim\p}[\text{HS}(\y_i, \y)] \\
         &= \argmax_{\y_i \in \{0,1\}^L} \left ({\sum_{\y_j \in \{0,1\}^L} \text{HS}(\y_i, \y_j)P(\y_j|\x)}\right ) \\
         &= \argmax_{\y_i \in \{0,1\}^L} \left ({\sum_{\y_j \in \{0,1\}^L}
           \frac{1}{L}\sum_{l=1}^{L}{ \big[ y_{il} = y_{jl} \big] }
           P(\y_j|\x)}\right ) \\
&= \argmax_{\y_i \in \{0,1\}^L} \left (
           \sum_{l=1}^{L}{ {\sum_{\y_j \in \{0,1\}^L} \big[ y_{il} = y_{jl} \big] }
           P(\y_j|\x)}\right ) \\
         &= \argmax_{\y_i \in \{0,1\}^L} \left (
           \sum_{l=1}^{L}{\mathbb{E}_{\rY\sim\p}\big[ y_{il} = y_{l} \big]}\right )
\end{align*}
The expected value of each label being 0 or 1, is positive, so we can compute the argmax of each element of the sum to obtain the prediction for each label:
\begin{align}
\hat{y}_l  &= \argmax_{z \in \{0,1\}} \left (\mathbb{E}_{\rY\sim\p}\big[ z = y_{l} \big]\right ) \text{ for each label } l
\end{align}
which relates to Eq.~\eqref{eq:tag}; showing that the chosen labelset consists of the mode for each label (marginal mode). It is worth remarking that the sum comes outside the expectation in the derivation of expected Hamming similarity.

In this paper we study candidate functions $f$ for measuring confidence in multi-label predictions, based on expected accuracy, for different accuracy metrics. More precisely, $f$ is based on $\p$ wrt $\mathcal{S}$ -- for a given $\x$. We assume that these components are all available and determined. 

\begin{figure}[h!]
  \begin{center}
	  \subfloat[][]{
		\includegraphics[scale=0.31]{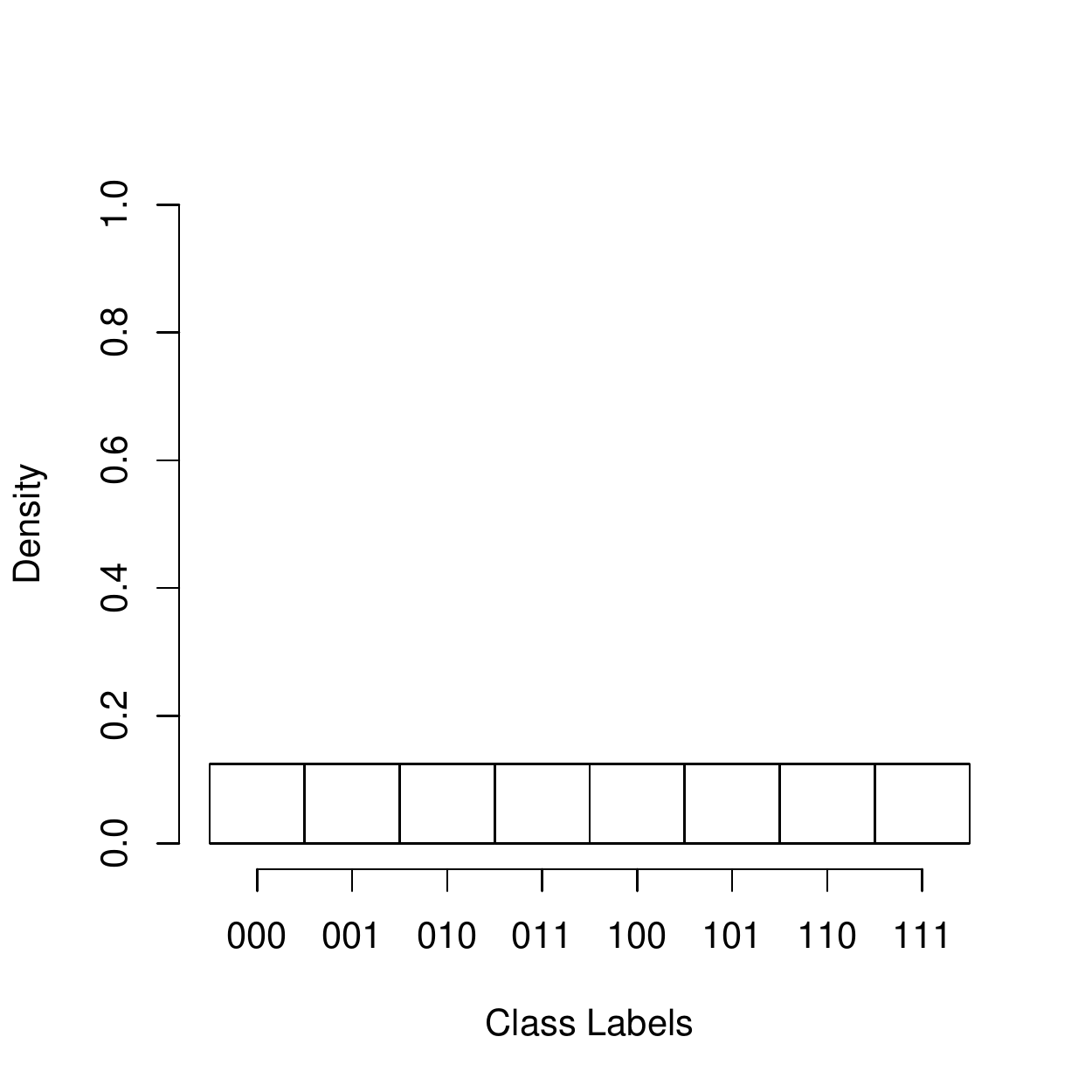}
		\label{fig:dist-example.a}
	  }
	  \subfloat[][]{
		\includegraphics[scale=0.31]{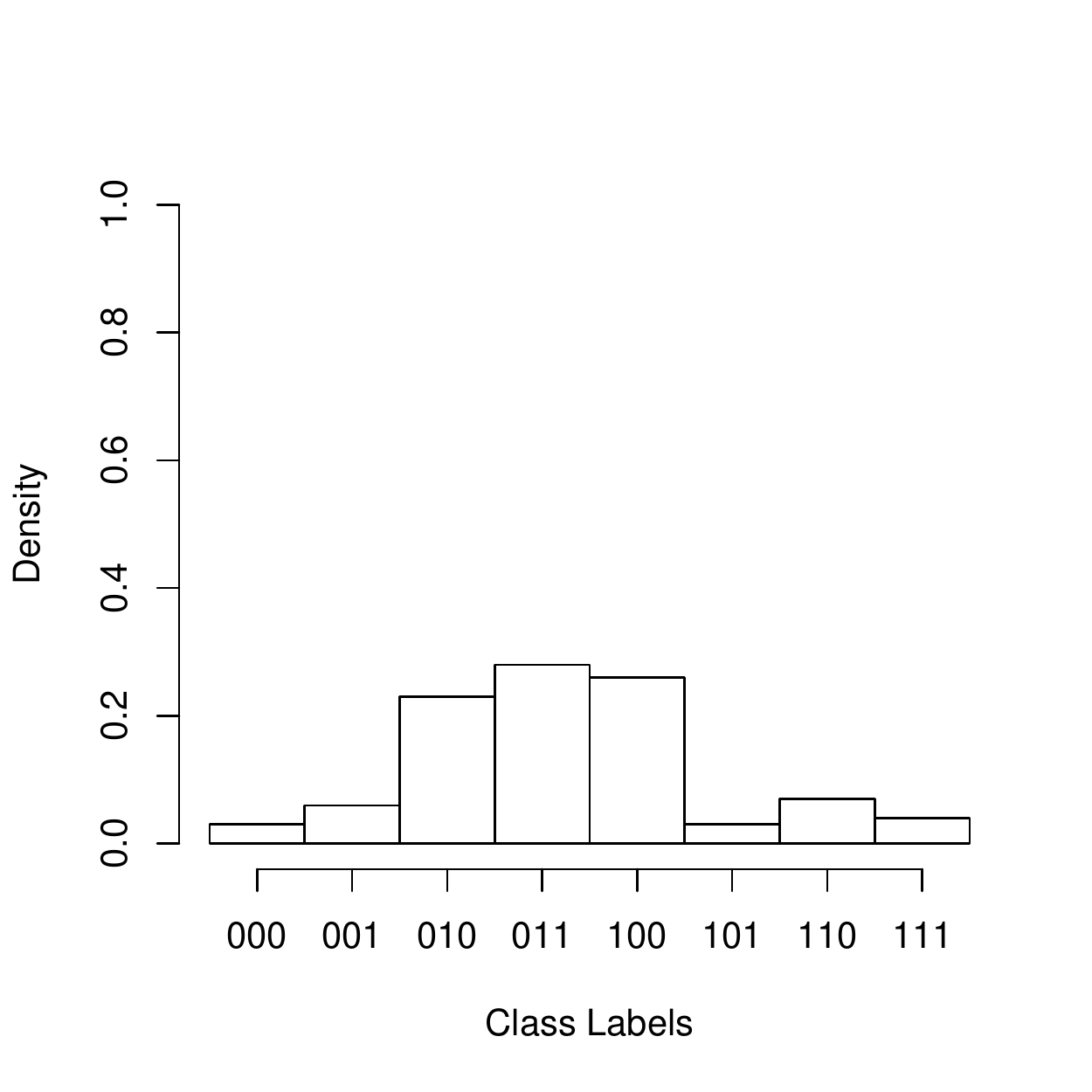}
		\label{fig:dist-example.b}
	  }
	  \subfloat[][]{
		\includegraphics[scale=0.31]{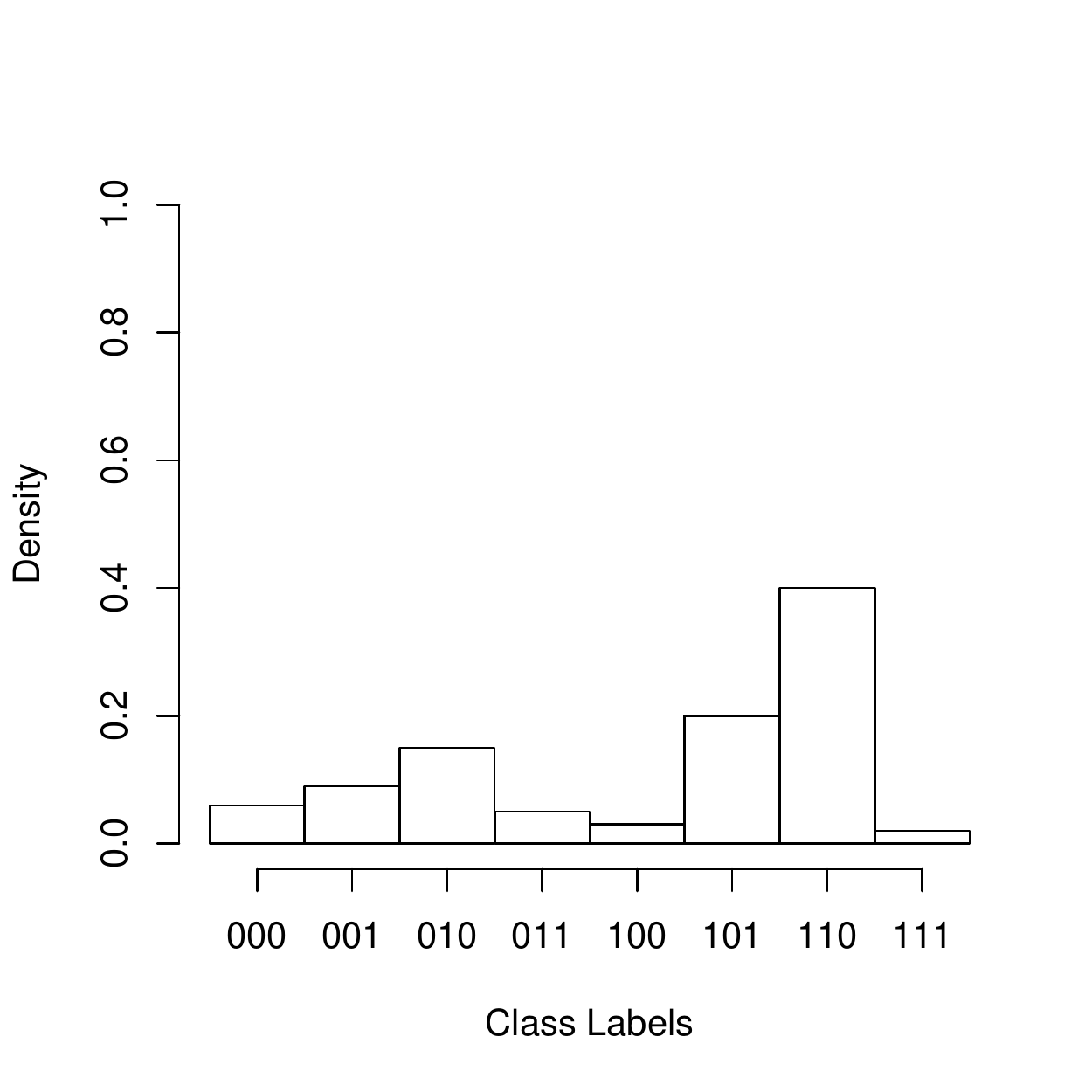}
	  }\\
	  \subfloat[][]{
		\includegraphics[scale=0.31]{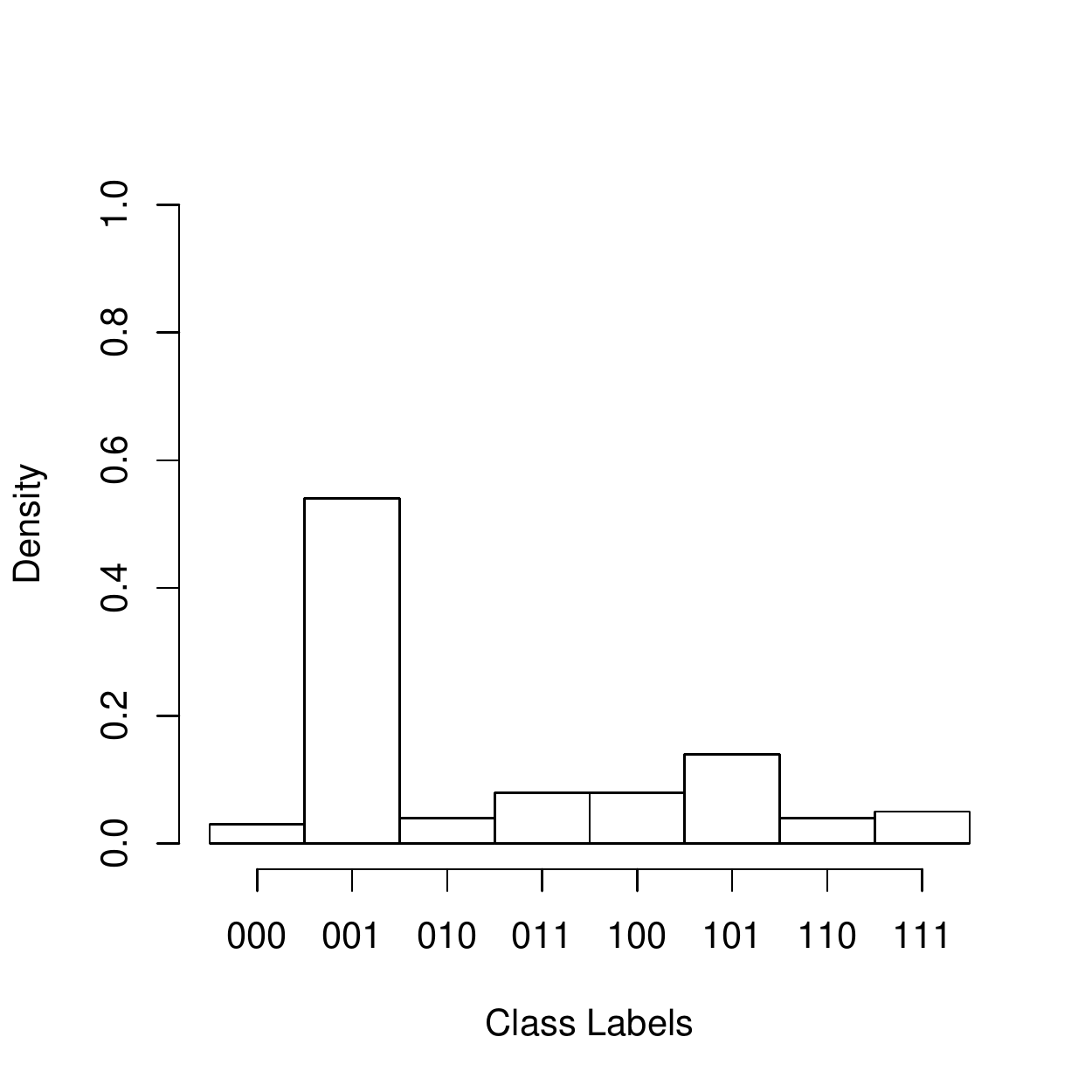}
	  }
	  \subfloat[][]{
		 \includegraphics[scale=0.31]{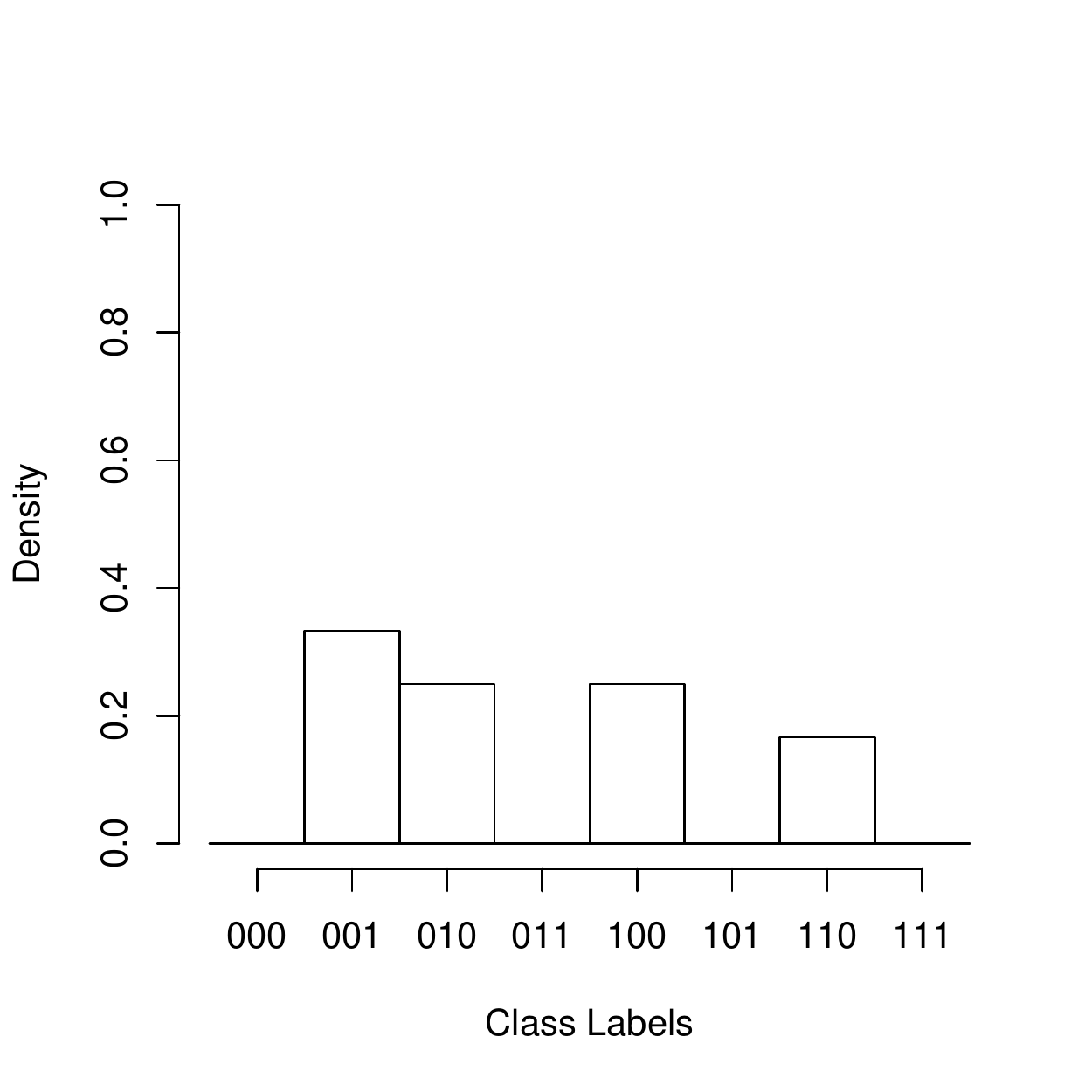}
		\label{fig:dist-example.e}
	  }
	  \subfloat[][]{
		\includegraphics[scale=0.31]{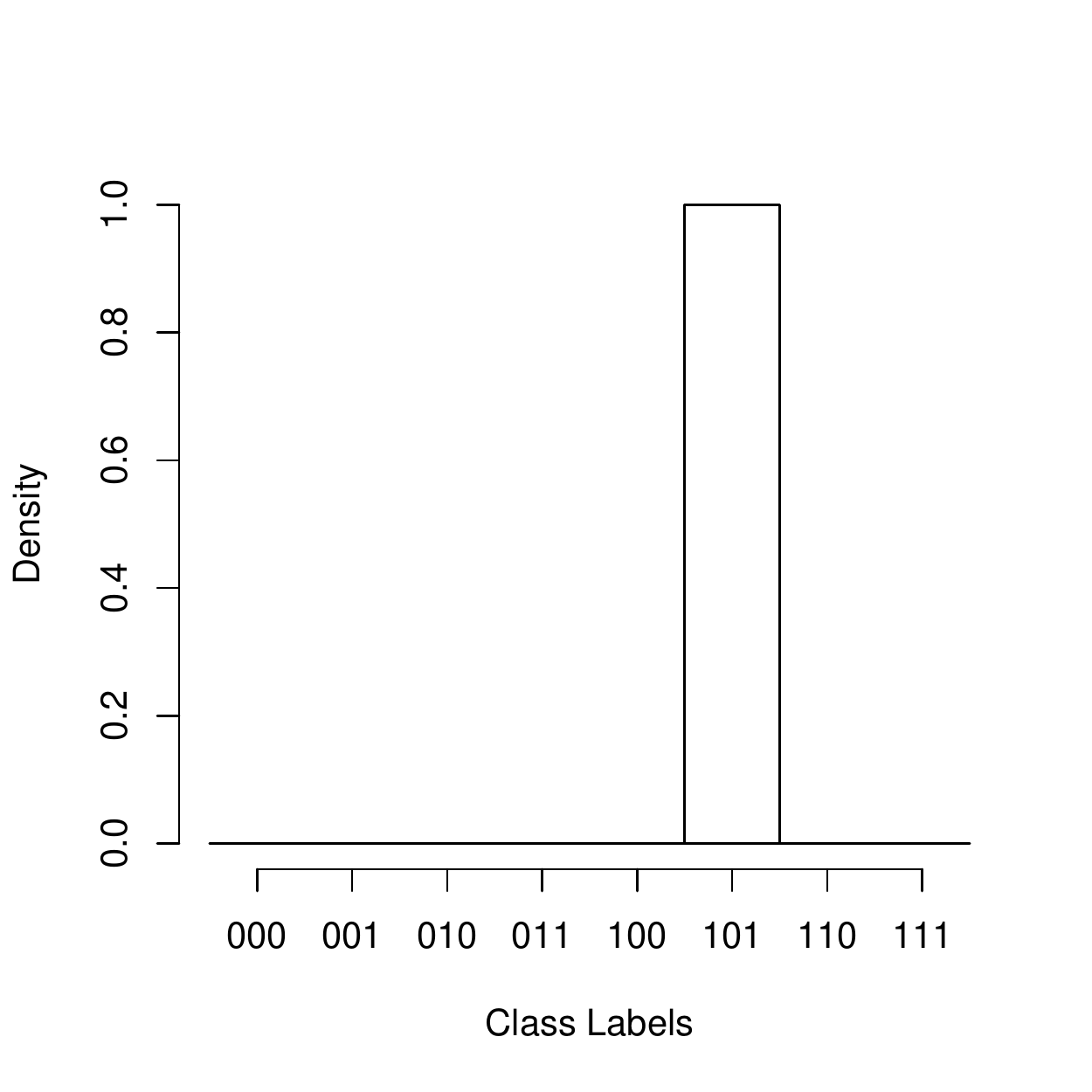}
		\label{fig:dist-example.f}
	  }
  \end{center}
	\caption{Six possible multi-label distributions $P(\y|\x)$ over three
    labels for a hypothetical given $\x$. It is clear that the top left distribution should provide
	the lowest expected accuracy (and correspondingly, the greatest uncertainty),
    and the bottom right distribution should provide high expected accuracy and the least uncertaintly regarding clasification, particularly with regard to exact match similarity,
    but it is not clear what level of expected accuracy we should provide to
    the remaining distributions. Distribution \fref{fig:dist-example.e} replicates graphically the distribution given in 
  \Tab{tab:example}.}
  \label{fig:dist-example}
\end{figure}

To illustrate with an example: Figure \ref{fig:dist-example} shows six hypothetical distributions $\p$ over the powerset of three labels, conditioned on some hypothetical instances $\x$. For the prediction, we could provide $\ymode$, the mode labelset of $\p$ known to maximize exact match similarity. For confidence in this prediction we could return $P(\ymode|\x)$. Intuitively we would expect that distribution \fref{fig:dist-example.a} results in a low expected accuracy since all labelset combinations are
equally unlikely, whereas distribution \fref{fig:dist-example.f} should result in high expected accuracy; yet it is not
obvious how we should measure the expected accuracy from the remaining
distributions. The goal of this paper is to formalise and test such intuitions, and provide a function $f$ with which to do so.

\section{Related Work}
\label{sec:related}

As we already mentioned in \Sec{sec:pml}, in the multi-label context it is not obvious as in single-label classification where one may simply look at returning a label along with its associated confidence, since many labels are involved. In this section we look at related methods for making decisions under uncertainty in context of classification and, and in particular to multi-label classification; in reflection of a posterior distribution $\p$.

\subsection{Thresholding}

A simple approach to gain trust in a probabilistic method is to calibrate a threshold according to desired confidence. For example, to only accept a label (or -- labelset -- in the multi-label context) if the probability it is relevant is greater than $0.9$ (or some other ad-hoc value). Thresholding has been considered in, e.g., \cite{MCut}.

This can help reduce false positives, but it does not cover the case where also \emph{not} applying the labelset (combination of labels) also carries significant risk; it encourages a large number of false negatives. It is also not clear how to set a threshold for label\emph{sets} for multi-label classification, since it is rare for any given combination of labels $\y_i$ to carry a high value in $P(\y_i|\x)$ (one may look at the examples in \Fig{fig:dist-example}).

\subsection{Ranking} 

Many authors, rather than attempting an explicit separation between relevant and irrelevant labels, have instead looked at \emph{ranking} labels \cite{UnifiedMultilabelLoss,ExtremeLoss}. An intrinsic label-wise ranking is readily available via their marginal scores $P(y_j|\x)$ (recall the example from \Tab{tab:example}, which corresponds to two equivalent best rankings: $(1,2,3) \equiv (2,1,3)$ from best to worst, wrt label index). 

Our problem is that we want to predict the confidence/accuracy of the instance, i.e., of the labelset \emph{combination}. And in this article we look at ranking but for investigating the relationship between score functions and expected accuracy; details and examples in \Sec{sec:order_equivalence}. This is different from using the ranking as a direct means to gauge relative label-wise confidence without a threshold.

\subsection{Reject Option} 

Classification with a \textit{reject option} (also known as abstention), has been an active research area \cite{bartlett2008classification,yuan2010classification,grandvalet2009support}. This is a natural extension of simple thresholding to express risk on positive (relevance) \emph{and} negative (not-relevant) labelling of a particular instance. Adding a reject option to a binary problem in this way adds a third state in which the classifier declares that it is not providing a prediction (i.e., it abstains). The three states are defined by partitioning the decision space into three regions ($\yp = 1$ true, $\yp = 0$ false, and reject), instead of the usual two partitions for binary classification. 

A new and interesting look at abstention in the context of multi-label classification 
is given in \cite{EykeRejectOptionNew}, building on  
\citet{pillai2013multi} where the rejection region was defined for each label using an upper an lower bound, and computed
using a cost function based on a predefined cost for rejection. In the multi-labelsetting one can consider partial abstention \cite{partialReject}, where a model may deliver predictions for a subset of class labels. 

In this work, we do not target this specific reject option, but rather the expression of confidence associated to a given labelset prediction.

\subsection{Cautious Prediction} 

One may consider being \emph{cautious} (or, equivalently, \emph{imprecise}) about predictions when uncertainty is high, and provide a set of possible labellings in the form of multiple predictions or posterior distributions $\{\p^m\}_{m=1}^M$ rather than a single one (the index $m$ is for an entire distribution); and rather than simply accepting the most likely labelset, or refusing to make a decision. In other words, the uncertainty for a given prediction is described by this \textit{credal set}.   These are also known also as \textit{distributionally robust} models, and they have recently been investigated in the multi-label case \cite{Yonatan,Sebastien}. 
However, the credal set itself does not answer the question we tackle in this study: how to gauge the potential accuracy of a given labelset prediction/combination $\y$ in reference to a given distribution/model $\p$.

\subsection{Skeptical Inference}

A further step beyond cautious prediction is to be \textit{skeptical} in the sense of considering as valid only those inferences that are true for every distribution within the credal set. For example (in the binary case) if the credal set involves $M=2$ distributions parametrized by $\{\theta_1,\theta_2\} = \{0.2,0.7\}$ (where $(P^m(Y=1|\x)) \Leftrightarrow \theta_m$ and thus $P^m(Y=0|\x) = 1- \theta_m$); then a positive label would \emph{not} be assigned, since both values are below a cut-off of $0.5$. The same process can be used on labelsets, for an appropriate cut-off, but then this produces connections to the threshold and reject options discussed above.

\subsection{Bayesian Inference}

Bayesian inference is a huge and well-established area of machine learning \cite{Barber} where one aims to obtain a posterior $P(\theta | \y, \mathbf{X})$ to model the uncertainty over the \emph{parameters} $\theta$ of the classifier. In general, a number of techniques, such as Monte Carlo methods, are needed to deal with obtaining such a model. Bayesian methods are popular particularly where human expert-knowledge can be taken into account into designing the prior distribution over parameters $P(\theta)$. 

The predictive posterior (in Bayesian terminology, in the multi-label case), is still $\p$ (as we have considered so far). In Bayesian inference one is usually interested in returning this full distribution as a guide to uncertainty, but a Bayes classifier (having to make a point-wise estimate/classification) under $0/1$ loss will provide a MAP estimate as per \Eq{eq:MAP}. 

There are numerous other points of overlap. For example, buried in the literature of particle filters (i.e., sequential important sampling; recursive Bayesian estimation; see, e.g., \cite{Djuric03}) in the signal processing community one finds the need to measure the \emph{degeneracy} of a distribution (the distribution represented by $L$ weighted particles; specifically, in order to carry out a re-sampling step). This is connected to what we want to achieve in the context of this work: degeneracy refers to a distribution which is flat, such as the one in \Fig{fig:dist-example.a}. Measuring this kind of degeneracy is in fact what we intend to do (albeit, for a different purpose). A typical choice in particle filters is the effective sample size
\[
	\textsf{ESS} = \left[\sum_{i=1}^n P(\y_i)^2\right]^{-1}
\]
(we have adapted the notation) which is along the vane of what we study in the following section (indeed, ESS is essentially a form of collision entropy without the $\log$); as we investigate the utility of the using the
multi-label distribution for each instance to estimate the expected accuracy of the predicted labelset, where expected accuracy is measured with respect to the chosen similarity function. A side remark: particle filters provide models for $P(\y|\x)$ where $\y \in \mathbb{R}^L$, unlike in our case; where $\y \in \{0,1\}^L$; a related treatment for multi-output regression is given in \cite{PRC}.

\section{Expected Accuracy Candidates for Multi-Label Classification}
\label{sec:acceptance}

We aim to use the posterior distribution $\p$ to provide a score representative of expected accuracy, 
	\begin{equation}
			\label{eq:expected_accuracy_fn}
f\left(\p\right) \in [0,1]
	\end{equation}
or more specifically $f(\x)$, where a value close to $1$ indicates that we expect to achieve high accuracy by predicting $\ypred$ for $\x$ (using $\p$ as per \Eq{eq:ysim}) and a value close to $0$ indicates that this prediction $\ypred$ is almost certainly wrong. 

In other words, we hypothesise that the expected accuracy is a function of the
multi-label label powerset distribution, but it is not clear how to
map the distribution to an expected accuracy value (i.e., what is the appropriate $f$).

We approach this problem of determining $f$ using traditional scientific
method, where we propose a set of functions $\{f_s\}$ corresponding to statistics of the distribution, and examine if the statistics provided
by the candidate functions are associated to expected accuracy, when
taking into account the environmental conditions (such as the data,
and parameters used). If an association is found, it provides evidence that $f_\textsf{s}$ is a component of $f$.

In \Sec{sec:acceptance_functions}, 
we will examine seven functions of this categorical
distribution that may be useful for
estimating the expected accuracy, in \Eq{eq:expected_accuracy_fn}.

Then in \Sec{sec:order_equivalence} we discuss how to make use of these functions for evaluating accuracy in practical experimental settings.

\subsection{Candidate Expected Accuracy Functions}
\label{sec:acceptance_functions}

We desire an \emph{expected accuracy} function $f$ that, given a labelset
distribution $\p$, returns a measurement of the distribution indicating the expected
accuracy of the prediction for a given instance.
There are two conditions that the
function must satisfy:
\begin{enumerate}
	\item $f=0$ if all labelsets have \label{cond:1} equal probability (e.g., in \Fig{fig:dist-example.a}).
	\item $f=1$ if the distribution \label{cond:2} contains one labelset with probability 1 and all others probability 0 (e.g., in \Fig{fig:dist-example.f}).
\end{enumerate}

In the following we examine candidate accuracy functions $f$ that meet these constraints and also provide suitable values $f \in (0,1)$ for ``in-between'' distributions (such as the four other distributions in Figure \ref{fig:dist-example}). 

We
will examine these functions in detail to determine which provide
values that are most closely associated to the expected
  accuracy for multi-label prediction using a given similarity metric.

\subsubsection{High probability (HP)}

The de facto standard for measuring confidence of multi-label
predictions due to its use in other forms of classification and
minimization of the Exact Match metric:
\begin{align}
  H_{\text{HP}} = {\max_{\vec{y}_i \in \sB^{L}} P(\vec{y}_{i})}
\end{align}
It is simply the probability of the labelset that has the greatest
probability (mode).  To remove the dependence of this function on the label
set size, we normalise the function so that the maximum value of 1 is
given when all but one item has zero probability, and minimum value of
0 is given when all items have equal probability;
\begin{align}
  C_{\text{HP}} = {\max_{\vec{y}_i \in \sB^{L}} 
  \frac{2^L P(\vec{y}_{i}) - 1}{2^L - 1}}
\end{align}
where $L$ is the number of labels. For example, for 3 labels, a
uniform distribution of $P(\y_i)=\frac{1}{8}$ (since there are 8
possible combinations) gives $C_{\text{HP}} = \frac{2^3 \cdot
  \frac{1}{8} - 1}{2^3 - 1} = 0$; the lowest expected accuracy. Since this function is the most commonly used, it is used as the baseline in each experiment.

\subsubsection{Top Gap (TG)}

HP only examines the mode of the multi-label distribution. However, it
may well be that the probability for other labelsets are very close
to this maximum value, in which case we can be less confident about a
correct prediction, as exemplified in \Fig{fig:dist-example.b}.

We also consider as a candidate the difference in
probability of the most probable label combination and the second most
probable label combination.  We call this function \textit{Top Gap}
(TG), which is given as:
\begin{align}
C_{\text{TG}} = \Big[ \max_{\vec{y}_i \in \sB^{L}} P(\vec{y}_{i}) \Big] 
	              - \Big[ \max_{\vec{y}_i \in \{\sB^{L} \setminus \ymode \}} P(\vec{y}_{i}) \Big] \\
				  = P(\ymode) - \Big[ \max_{\vec{y}_i \in \{\sB^{L} \setminus \ymode \}} P(\vec{y}_{i}) \Big]
\end{align}
The intuition behind the function is that a labelset distribution
with a large TG implies greater confidence and hence higher expected
accuracy, since one labelset has a much greater probability then the others.

If $C_{\text{TG}}$ is small, then there is little difference between
the top two probabilities, meaning that there is uncertainty as to
which of the associated label combinations is the best choice.  If
$C_{\text{TG}}$ is large, then there is a large difference in
probability between the most probable label combination and all of the
other label combinations, meaning that there is a clear choice from
the probability distribution.

This function is already appropriately limited between 0 and 1, therefore no normalisation is needed.

\subsubsection{Shannon Entropy (SE)}

Shannon entropy is a measure of the information content from a
channel, given the distribution of symbols transmitted through the
channel. The greater the uncertainty of the symbols, the greater the
information content. Using our notation, the entropy of the
multi-label distribution is:
\begin{align}
  H_{\text{SE}} = -{\sum_{\vec{y}_i \in \sB^{L}} P(\vec{y}_{i})\log{\left (P(\vec{y}_{i}) \right )}}
\end{align}
which measures the uncertainty provided by the labelset distribution
\citep{shannon1948}
($H = 0$ means no uncertainty), where $0\times\log{(0)} = 0$. The
range of $H$ depends on the number of labelset combinations in our
multi-label problem. To adjust the range to $[0,1]$, we can use
normalised entropy:
\begin{align}
  H_{\text{SE}}^{\star} = -{\sum_{\vec{y}_i \in \sB^{L}} P(\vec{y}_{i})\frac{\log{\left (P(\vec{y}_{i}) \right )}}{\log{(2^L)}}}
  = -{\sum_{\vec{y}_i \in \sB^{L}} P(\vec{y}_{i})\log_{2^L}\!{\left (P(\vec{y}_{i}) \right )}}
\end{align}
where $H_{\text{SE}}^{\star} = 1$ is provided when the probability of all class
combinations are equal \citep{ParkS15}. 
Note that normalised entropy is independent of the entropy log
base, since the normalisation converts the base to $2^L$.

To obtain an expected accuracy candidate, we subtract
the entropy score from 1 to obtain
\begin{align}
  C_{\text{SE}} = 1 + {\sum_{\vec{y}_i \in \sB^{L}} P(\vec{y}_{i})\log_{2^L}\!{\left (P(\vec{y}_{i}) \right )}}
\end{align}
where $C_{\text{SE}} \in [0,1]$ is a measure of expected accuracy
using Shannon Entropy ($C_{\text{SE}} = 0$ implies no confidence and
low expected accuracy, while $C_{\text{SE}} = 1$ implies full
confidence and high expected accuracy for the prediction.

\subsubsection{Collision Entropy (CE)}

Collision entropy (CE) is a commonly used form of R\'enyi entropy, with
parameter $\alpha = 2$.
\begin{align}
  H_{\text{CE}} = -\log{\sum_{\vec{y}_i \in \sB^{L}} {P(\vec{y}_{i})}^2}
\end{align}
Collision entropy can be normalised to provide an expected accuracy candidate,
giving a score of zero when the distribution is Uniform, and a score
of 1 when all but one of the label combinations are zero:
\begin{align}
  \label{eq:ce}
  C_{\text{CE}} = 1 + \log_{2^L}{\sum_{\vec{y}_i \in \sB^{L}} {P(\vec{y}_{i})}^2}
\end{align}

\subsubsection{Min Entropy (ME)}

The amount of information in (number of bits required to represent) an
object is measured as the log of the reciprocal of the object's
probability. Shannon Entropy is the expected value of this information
measure across all possible objects. Rather than measuring the
expected value (mean), we can measure the minimum information across
all objects using Min Entropy (ME):
\begin{align}
  H_{\text{ME}} = -\log{\max_{\vec{y}_i \in \sB^{L}} P(\vec{y}_{i})}
  \label{eq:min_entropy}
\end{align}
This entropy measurement can be normalised to obtain a potential
measure of expected accuracy:
\begin{align}
  C_{\text{ME}} = 1 + \log_{2^L}{\max_{\vec{y}_i \in \sB^{L}} P(\vec{y}_{i})}
  \label{eq:min_entropy_norm}
\end{align}
where $C_{\text{ME}} \in [0,1]$, $C_{\text{ME}} = 0$ implies low
expected accuracy, while $C_{\text{ME}} = 1$
implies high expected accuracy (full confidence in the prediction).

\subsubsection{Gini Impurity}

Gini impurity (GI) is a common measure of impurity, used to determine the
branching variables in a decision tree. Gini impurity increases as the
distribution approaches Uniform and provides a score of zero for
distributions containing probabilities of zero for all but one item.
Gini impurity is given as:
\begin{align}
  H_{\text{GI}} = {\sum_{\vec{y}_i \in \sB^{L}} P(\vec{y}_{i}){\left
        (1 - P(\vec{y}_{i}) \right )}}
\end{align}
A normalised from of Gini impurity provides us with an potential
measure of expected accuracy:
\begin{align}
  C_{\text{GI}} = 1 - {\sum_{\vec{y}_i \in \sB^{L}} \frac{P(\vec{y}_{i}){\left
        (1 - P(\vec{y}_{i}) \right )}}{1 - 2^{-L}}}
\end{align}
where $C_{\text{GI}} = 0$ when all labels have the same probability,
and $C_{\text{GI}} = 1$ when all but one item have zero probability.

\subsubsection{Chi-squared Statistic}

The Chi-squared statistic (CS) is a measure of difference between the set
of expected and observed sample frequencies from a multinomial
distribution. The statistic follows a Chi-squared distribution, if the
observed frequencies come from the expected distribution.  For our
use, we compare probabilities rather than frequencies, and we set the
expected distribution to be Uniform, so the further the computed
distribution is from Uniform, the greater the score.
\begin{align}
  H_{\text{CS}} = {\sum_{\vec{y}_i \in \sB^{L}} \frac{{\left (P(\vec{y}_{i}) - 2^{-L} \right )}^2}{2^{-L}}}
\end{align}
The normalised Chi-square statistic ensures a maximum of 1 when all
but one of the items have zero probability, and a score of 0 when all
items have equal probability.
\begin{align}
  C_{\text{CS}} = {\sum_{\vec{y}_i \in \sB^{L}} \frac{{\left (P(\vec{y}_{i}) - 2^{-L} \right )}^2}{1 - 2^{-L}}}
\end{align}

\begin{figure}
  \centering
  \includegraphics[width=0.32\linewidth]{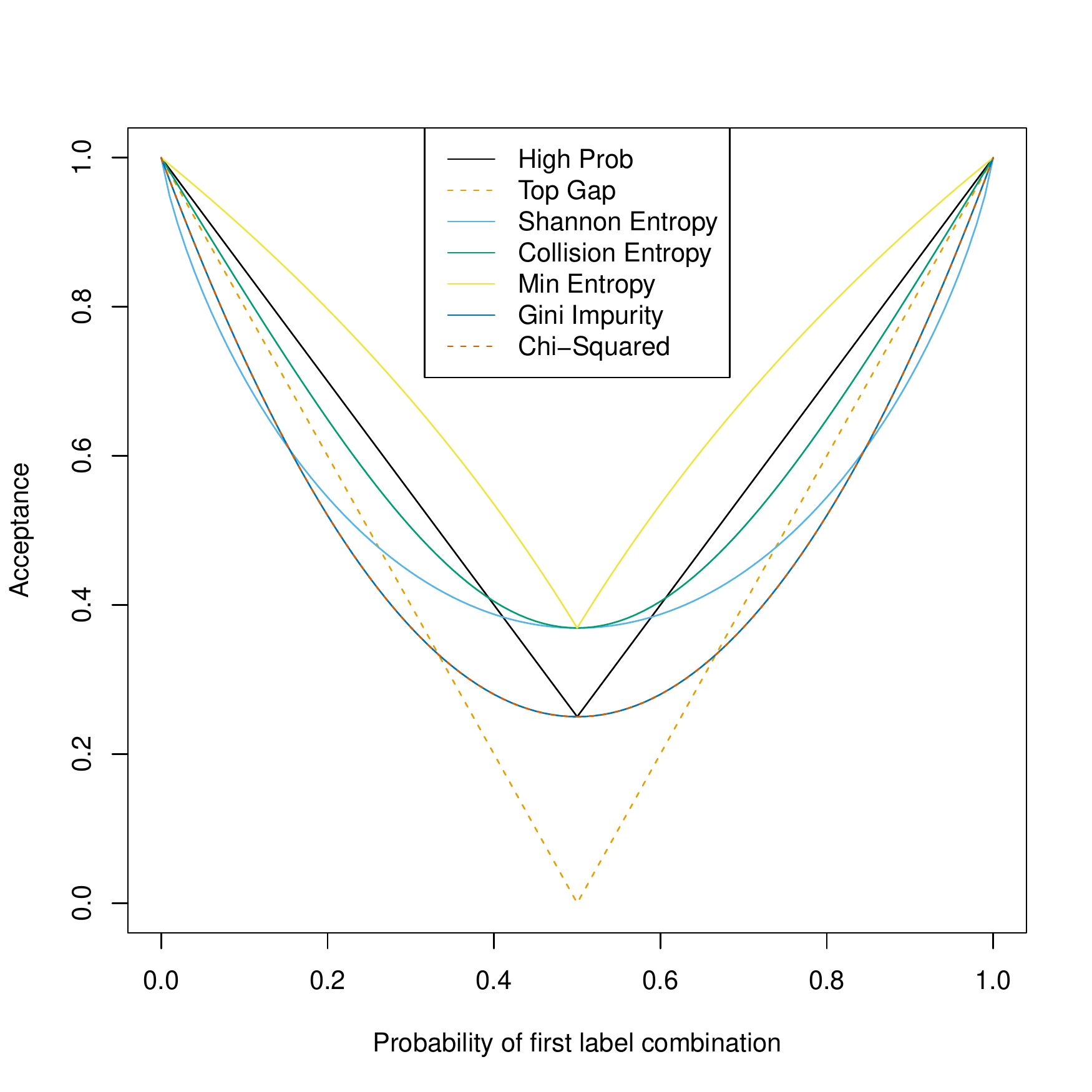}
  \includegraphics[width=0.32\linewidth]{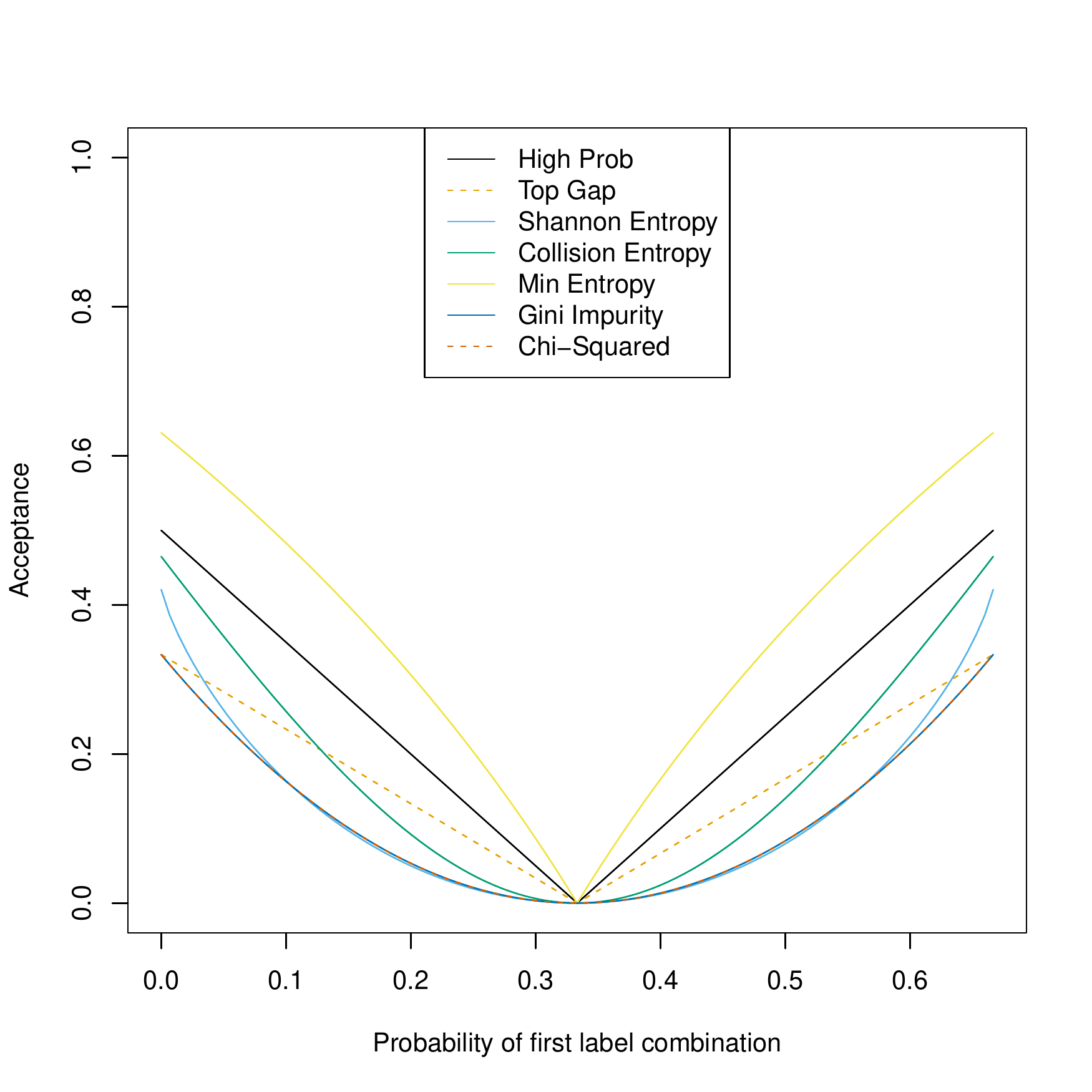}
  \includegraphics[width=0.32\linewidth]{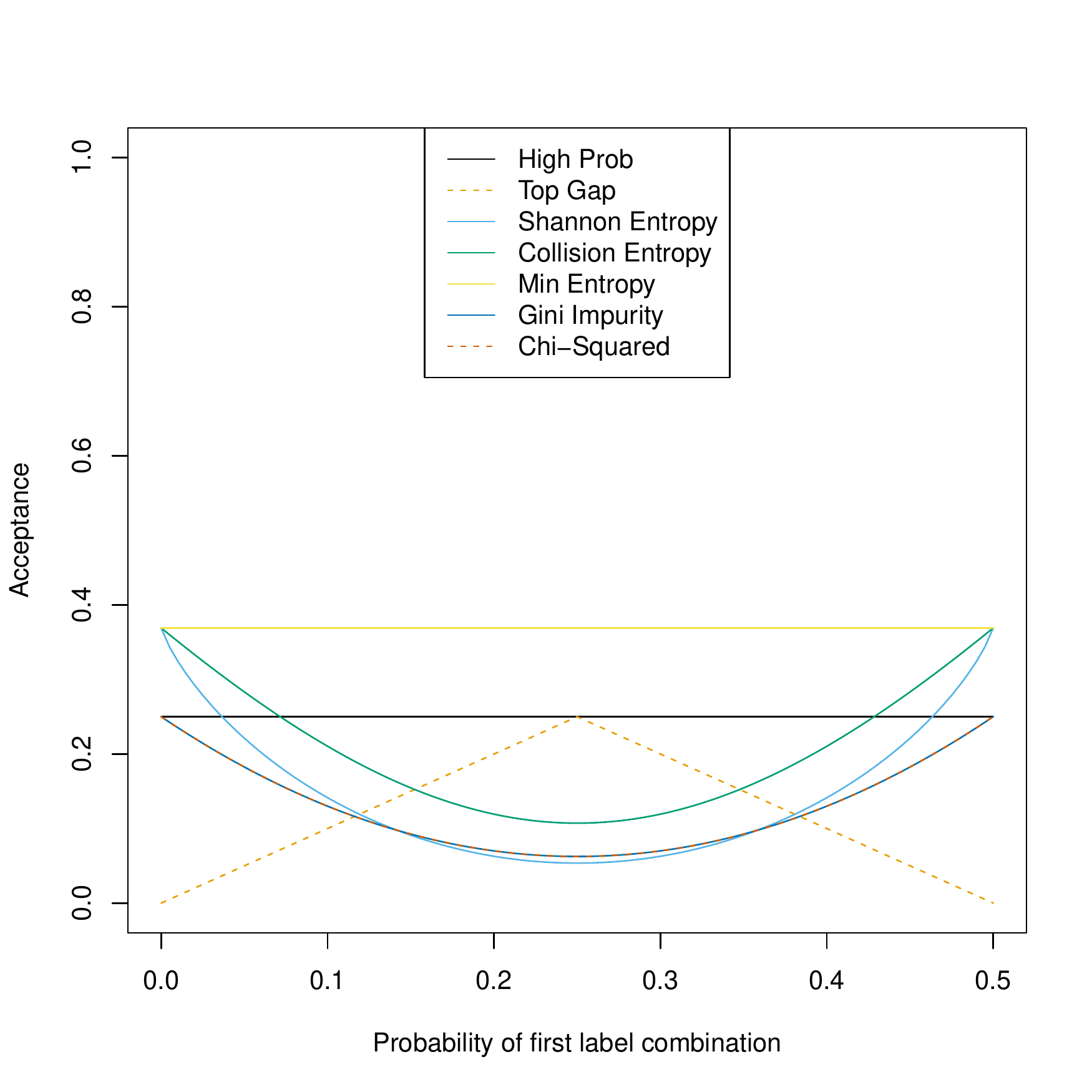}
	\caption{A comparison of the seven candidate accuracy functions 
    over a probability distribution with three elements. The $x$ axis
    shows the probability associated to the first of the three
    elements, the probability of the third element is 0 for the left
    plot, 1/3 for the middle plot and 1/2 for the right plot; the $y$
    axis shows the expected accuracy provided by the candidates. All functions are
    normalised to provide a value of 1 when all but one item has
    probability zero, and a value of zero when all items have equal
    probability.}
\label{fig:compareCandidates}
\end{figure}

A comparison of the behaviour of the seven candidate expected accuracy
functions $C_{\text{HP}}$, $C_{\text{TG}}$, $C_{\text{SE}}$,
$C_{\text{CE}}$, $C_{\text{ME}}$, $C_{\text{GI}}$ and $C_{\text{CS}}$,
on a multinomial distribution of size three, is shown in Figure
\ref{fig:compareCandidates}. The left figure shows the effect of
changing the first and second probabilities while holding the third to
0. The normalisation has forced each of the functions to have a 
 score of 1 when the first or second probability is 1
(meaning the remaining probabilities are 0). For the middle plot, the
third probability is $1/3$. We see that all functions have a
value of 0 when all probabilities are equal (all $1/3$). The plot on
the right has the third probability set to $1/2$. The three plots show
that High Prob and Top Gap are linear except at the cusp points, Min
Entropy is concave, while the remaining functions are convex. All
functions show the same decrease then increase shape, except for Top
Gap in the right plot, which increases then decreases (due to it only
observing the difference between the top two probabilities). We also
find that High Prob and Min Entropy are flat in the right plot, since
they only depend on the greatest probability, which does not change
for this plot. Finally, we find that Chi-Squared and Gini Impurity
have provided equal expected accuracy for all probability values in all
plots.

\subsection{Relative and Absolute Association}
\label{sec:order_equivalence}

The candidate expected-accuracy functions described above
(\Sec{sec:acceptance_functions}; generically stated in
\Eq{eq:expected_accuracy_fn}) provide a measure $f$ for the labelset
distribution of each instance. This score provided by $f$ has the
potential to be used in practice to determine the suitability of the
model for the given instance $\x$ (i.e., to estimate the expected accuracy of its prediction for that instance), with higher values (closer to $1$) implying that the model is more suitable. For the score to be useful it must have some association to the actual
accuracy of the prediction. Therefore, the candidate functions will be
assessed based on their association to the true expected accuracy.

We discuss two levels of expected accuracy association: a weaker form
of association that we call \textit{relative association} and stronger form
that we call \textit{absolute association}.

\medskip

Let $\D = \{(\x_n,\ypred_n)\}_{n=1}^N$ be a data set of $N$ \emph{test} instances alongside their predictions. Let $r_n \in \{1,\ldots,N\}$ indicate the ranking of the $n$-th instance, such that $r(f) := r_1,\ldots,r_N$ is a permutation of $1,\ldots,N$, corresponding to rank \emph{when scored by} candidate function $f$. In this context, $r_1(f)$ is the index of the instance providing highest score $f(\P(\y|\x_{r_1}))$, and so on. Likewise, let $r_1(E)$ correspond to the index of the instance giving highest expected accuracy (as per \Eq{eq:Exp}), i.e., $E(\ypred_{r_1})$ is the highest value, and $E(\ypred_n)$ is the lowest. Recall that the expected accuracy is with respect to a similarity function of interest, $\mathcal{S}$. 

\medskip

\textbf{Relative Association:} A candidate function $f$ provides a \emph{relative association} to
expected accuracy (with respect to a given similarity function) if the ordering of the instances $r_1(f),\ldots,r_N(f)$ provides a close match (e.g., monotonically increasing) to the ordering
of the instances provided by their expected accuracy $r_1(E),\ldots,r_N(E)$. In other words we measure
\[
	\textsf{RelativeAssociation}(r(f),r(E))
\]
where $f$ and $E$ are both associated with some similarity function $\mathcal{S}$. The association will not necessarily be the same for different $\mathcal{S}$. 

An example is given in \Tab{tab:eg_assoc}. 

\begin{table}[h!]
	\caption{An example for $N=3$ instances, $\x_1,\x_2,\x_3$, supposing
          $f := f_{\textsf{ME}}$ (i.e, using Min Entropy) and $E
          := E_{\textsf{HS}}$ (i.e., using Hamming Similarity). In
          relative association we compare the ranks (columns $r(E)$ vs
          $r(f)$). In absolute association, we compare columns $E$ and
          $f$. In experiments we do this for different classes of $f$
		  and $E$; and different similarity metrics $\mathcal{S}$.}
                \begin{center}
                  \begin{tabular}{rcccccccc}
    \toprule
		$i$ & $\x_i$ & $\ypred_i = [\hat Y_1$ & $\hat Y_2$ & $\hat Y_3 ]$ & $E$             & $f$& $r(E)$& $r(f)$ \\
    \midrule                                                            
		1 & $x_1$ &	$[ 0$   & $0$   & $0 ]$   & 0.6  & 0.2   & 1   & 3 \\
		2 & $x_2$ &$[ 0$   & $0$   & $1 ]$   & 0.5  & 0.8   & 2   & 2\\
		3 & $x_3$ &$[ 0$   & $1$   & $0 ]$   & 0.5           & 3.1   & 2   & 1\\
\bottomrule
                  \end{tabular}
                  \end{center}
	\label{tab:eg_assoc}
\end{table}

\textbf{Absolute Association:} 
A candidate function $f$ provides an \emph{absolute association} to
expected accuracy if the scores provided by each are associated. In other words we measure 
\[
	\textsf{AbsoluteAssociation}(f,E) 
\]
 
Absolute association is stronger, since it implies that we are able to
predict the expected accuracy from the candidate
function. Whereas relative association only provides us with which
instances have a greater expected accuracy than others (suggesting
that there is a monotonic relationship).

We begin by examining relative association and then follow by
examining absolute association to expected accuracy.

When measuring relative expected accuracy, we examine the order (ranking) of
the scores, not the values of the scores, therefore some of the
candidate expected accuracy functions become equivalent.
We find that both CE and Gini impurity, and Min entropy and High probability provide equivalent orderings/rankings 

With regard to min entropy (given in \Eq{eq:min_entropy_norm}),
the terms $1 + y$ and $\log_{2^L}(y)$ are monotonic functions which do not effect the order of $y$. Therefore the ordering  provided by $C_{\text{ME}}$ is equivalent to the ordering provided by $C_{\text{HP}}$.

Gini impurity is given as:
\begin{align}
  C_{\text{GI}} &= 1 - {\sum_{\vec{y}_i \in \sB^{L}} \frac{P(\vec{y}_{i}){\left
        (1 - P(\vec{y}_{i}) \right )}}{1 - 2^{-L}}} \\
         &= 1 - {\sum_{\vec{y}_i \in \sB^{L}} \frac{P(\vec{y}_{i})
            - P(\vec{y}_{i})^2 }{1 - 2^{-L}}} \\
         &= 1 - \frac{{\sum_{\vec{y}_i \in \sB^{L}} P(\vec{y}_{i})}}{1 - 2^{-L}} + \frac{{\sum_{\vec{y}_i \in \sB^{L}} P(\vec{y}_{i})^2}}{1 - 2^{-L}} \\
         &= 1 - \frac{1}{1 - 2^{-L}} + \frac{{\sum_{\vec{y}_i \in \sB^{L}} P(\vec{y}_{i})^2}}{1 - 2^{-L}} \\
         &= \frac{- 2^{-L}}{1 - 2^{-L}} + \frac{{\sum_{\vec{y}_i \in \sB^{L}} P(\vec{y}_{i})^2}}{1 - 2^{-L}}
\end{align}
which provides an equivalent ordering to
${\sum_{\vec{y}_i \in \sB^{L}} P(\vec{y}_{i})^2}$. By examining
equation \ref{eq:ce}, and again noting that $1 + \log_{2^L}(x)$ is
monotonic, we find that CE has an equivalent ordering
to ${\sum_{\vec{y}_i \in \sB^{L}} P(\vec{y}_{i})^2}$ and hence
also to Gini impurity.
Also
\begin{align}
  C_{\text{CS}} &= {\sum_{\vec{y}_i \in \sB^{L}} \frac{{\left (P(\vec{y}_{i}) - 2^{-L} \right )}^2}{1 - 2^{-L}}} \\
                &= {\sum_{\vec{y}_i \in \sB^{L}} \frac{P(\vec{y}_{i})^2 - 2^{1-L}P(\vec{y}_{i}) + 2^{-2L}}{1 - 2^{-L}}} \\
                &= - \frac{2^{1-L}}{1 - 2^{-L}} + \frac{2^{-L}}{1 - 2^{-L}} + {\sum_{\vec{y}_i \in \sB^{L}} \frac{P(\vec{y}_{i})^2}{1 - 2^{-L}}}  \\
                &= - \frac{2^{-L}}{1 - 2^{-L}} + {\sum_{\vec{y}_i \in \sB^{L}} \frac{P(\vec{y}_{i})^2}{1 - 2^{-L}}}  
\end{align}
provides an equivalent ordering to
${\sum_{\vec{y}_i \in \sB^{L}} P(\vec{y}_{i})^2}$ and is
identical to $C_{\text{GI}}$. Showing that $C_{\text{GI}}$,
$C_{\text{CE}}$ and $C_{\text{CS}}$ provide the same ordering.

Therefore, when analysing relative accuracy, we examine only HP, TG,
SE and CE. When analysing absolute accuracy, we examine HP, TG, SE,
CE, ME and CS.

\subsection{Bias in sample entropy}
\label{sec:entbias}

Estimating the expected accuracy of the multi-label distribution using
the seven candidates requires us to first compute the
probability of each labelset combination in the $2^L$ size categorical
distribution. Computing this distribution can be time consuming,
depending on the number of labels. For example, given $50$ labels, we
have a categorical distribution with $2^{50} = 1.1259\times 10^{15}$
items. If it takes a millisecond to compute the probability of one
item, it will take $35702$ years to compute the set of $2^{50}$
probabilities. To avoid this problem, we can either limit ourselves to
data with a smaller number of labels, or approximate the distribution.

Each multi-label distribution can be approximated by sampling from the
conditional distribution of each label, to obtain a sample from the
multi-label distribution. But unfortunately, sample estimates of
entropy are biased. Analysis has shown how to correct for such bias
\cite{schurmann2004bias}, but not for categorical distributions with a
huge number of categories. It is easy to see that the problem comes
from estimating the long tail probabilities. For example, using 50 labels, leading to a 
categorical distribution containing $1.1259\times 10^{15}$ items, 
a sample of size $1,000,000$ will miss many of the items in the tail
and underestimate their contribution to be $0$ ($0\times\log{0}$). 
Further research must be performed to identify how we can avoid or correct for the bias in sample entropy, but we will leave that for future work.

During our analysis, we want to control as many of the variables as
possible to ensure that our experimental results are meaningful and
not confounded with unknown effects.  We do not want this sample bias
to be a factor that influences this analysis, therefore, we will avoid
sampling and compute the whole distribution, but limit the data to
manageable sizes.

\section{Experimental Setup}
\label{sec:Experiments}

In this section we describe the experimental setup. Then, in \Sec{sec:relative} and \Sec{sec:absolute} we look at and discuss the experimental results with regard to relative and absolute expected accuracy, respectively.

An extensive experimental evaluation is conducted to investigate which
candidate functions (of those detailed in
\Sec{sec:acceptance_functions}) are associated to the expected
accuracy, and in which context (which kind of dataset, for what kind
of classification method, and under which evaluation metric).

We use three probabilistic methods involving independence (BR), some
dependence (CT) and full-dependence (ECC) modelling, described and referenced in
\Sec{sec:intro}; thus encompassing the
full range of possible dependence modelling of the labels. We use
three evaluation metrics -- Exact Match, Hamming Similarity and
Jaccard Similarity -- as described in \Sec{sec:sim}.

Datasets are listed in Table \ref{tab:data}; available from the MULAN\footnote{\url{http://mulan.sourceforge.net/datasets-mlc.html}}, MEKA\footnote{\url{https://sourceforge.net/projects/meka/files/Datasets/}}, and STARE \footnote{\url{http://www.ces.clemson.edu/~ahoover/stare/}} (for the \dataset{Stare} dataset) repositories.
Note again that we have selected data containing no more that 25 labels to avoid sampling bias (described in Section \ref{sec:entbias}). This is not an article about extreme multi-label classification (for reference, see, e.g., \cite{ExtremeML}, therefore we do not consider scalability a primary concern.

Thus, the variables involved in our experiment are: 
1) the candidate expected accuracy function, 2) the data set, 3) the
accuracy metric; and 4) the method of multi-label classification.

We investigate the utility of the candidate functions in two stages.
\begin{enumerate}
\item Relative accuracy identifies if the candidate has an association
  with expected accuracy.
\item Absolute accuracy identifies the potential for using the
  candidates estimate for measuring expected accuracy.
\end{enumerate}
We examine these in \Sec{sec:relative} and \Sec{sec:absolute}, respectively.

\begin{table}
	\centering
	\begin{tabular}{rrrrcccr}
	\toprule
	    		& $N$  &$L$ &	$M$		&LC &Type 		 \\           
	\midrule           
		\dataset{Emotion}       &593    &6     &72      &1.87          &audio   \\    \dataset{Scene}   	&2407	&6     &294		&1.07          &image	 \\     \dataset{Yeast}   	&2417 	&14   	&103	   	&4.24          &biology \\     \dataset{Stare}		&373 	&15      &44    &1.32          &medical	 \\     \dataset{Enron}   	&1702	&20 (53)    &1001    &3.38          &text	 \\     \dataset{Slashdot} 	&3782	&22    	&1079	&1.18          &text	 \\     \bottomrule
	\end{tabular}
	\caption{\label{tab:data} Data sets and associated statistics, where LC is \emph{label cardinality}: the average number of labels relevant to each example.}
\end{table}

\section{Relative Expected Accuracy}
\label{sec:relative}

A candidate function provides high relative expected accuracy if it is
weakly monotonically increasing with respect to the chosen evaluation
score, implying an association.
To examine if an expected accuracy candidate
provided a weak increasing monotonic relationship to the evaluation
scores, the Kendall's $\tau$ correlation between the accuracy and
candidate scores was computed for each combination of variables:
candidate function, multi-label method, data set and evaluation
function, giving 216 correlation results. These results are presented
in Appexdix \ref{sec:rawcor} using HP as the baseline.  The results
show that TG is generally the same or has lower correlation than HP,
but the for remaining candidate functions there are some cases that
have higher correlation, some the same and some worse.

According to these results, it is clear that the data, method and metric variables effect the
correlation, but it is not clear \textit{how} they effect the
correlation. To obtain insight into the effect of each variable on the
[accuracy $-$ candidate score] correlation, we provide a marginal
analysis for accuracy and robustness.

\stepcounter{AnalysisNumber}

\subsection{Analysis \theAnalysisNumber: Accuracy Analysis for Relative Accuracy}

Having found that there is merit in the candidate expected accuracy
functions other than HP in particular conditions, we now delve deeper and generalise the results. In particular, we examine
the marginal effect of each experimental variable (candidate function, dataset and method -- under each metric) on the
correlation. 

We use Fisher's $Z$ transformation
\cite{hotelling1953new} to provide variance stabilisation for the
correlation values to allow us to model the effect of each of the
variables on the correlation using a linear fixed effects model.
The results are presented in Table \ref{table:kcoefficients}.

\begin{table*}
\begin{center}
\begin{tabular}{l D{)}{)}{13)3} D{)}{)}{13)3} D{)}{)}{13)3} }
\toprule
 & \multicolumn{1}{c}{Exact} & \multicolumn{1}{c}{Jaccard} & \multicolumn{1}{c}{Hamming} \\
\midrule
\multicolumn{4}{l}{\textit{Expected Accuracy Function Bias (Baseline: High Prob)}} \\
\zspace{}Top Gap              & -0.058 \; (0.017)^{\star\star\star} & -0.070 \; (0.018)^{\star\star\star} & -0.073 \; (0.013)^{\star\star\star} \\
\zspace{}Shannon Entropy              & 0.008 \; (0.017)        & 0.050 \; (0.018)^{\star\star\star}  & 0.017 \; (0.013)        \\
\zspace{}Collision Entropy              & 0.012 \; (0.017)        & 0.030 \; (0.018)^{\star}    & 0.023 \; (0.013)^{\star}    \\[0.5em]
\textit{Data Bias (Baseline: Emotions)} \\
\zspace{}Scene             & 0.124 \; (0.021)^{\star\star\star}  & 0.094 \; (0.022)^{\star\star\star}  & 0.138 \; (0.016)^{\star\star\star}  \\
\zspace{}Yeast             & 0.109 \; (0.021)^{\star\star\star}  & 0.024 \; (0.022)        & 0.070 \; (0.016)^{\star\star\star}  \\
\zspace{}Stare             & -0.001 \; (0.021)       & -0.006 \; (0.022)       & 0.036 \; (0.016)^{\star\star}   \\
\zspace{}Enron             & 0.128 \; (0.021)^{\star\star\star}  & -0.293 \; (0.022)^{\star\star\star} & 0.157 \; (0.016)^{\star\star\star}  \\
\zspace{}Slashdot          & -0.050 \; (0.021)^{\star\star}  & -0.094 \; (0.022)^{\star\star\star} & -0.018 \; (0.016)       \\[0.5em]
\textit{Classifier Bias (Baseline: ECC)} \\
\zspace{}Independent & -0.069 \; (0.015)^{\star\star\star} & -0.085 \; (0.015)^{\star\star\star} & -0.074 \; (0.011)^{\star\star\star} \\
\zspace{}CT          & -0.020 \; (0.015)       & -0.027 \; (0.015)^{\star}   & -0.046 \; (0.011)^{\star\star\star} \\
\midrule
R$^2$                  & 0.751                   & 0.886                   & 0.838                   \\
Adj. R$^2$             & 0.710                   & 0.867                   & 0.812                   \\
Num. obs.              & 72                      & 72                      & 72                      \\
RMSE                   & 0.051                   & 0.053                   & 0.039                   \\
\bottomrule
\multicolumn{4}{l}{\scriptsize{$^{\star\star\star}p<0.01$, $^{\star\star}p<0.05$, $^\star p<0.1$}}
\end{tabular}
\caption{The expected change in variance stabilised Kendall's $\tau$
  correlation for the three variables (Expected Accuracy function, Data,
  Classifier) and standard error in parentheses, with respect to each
  variable's baseline, independent of the other variable values. A
  value of 0 represents no change in the correlation when changing
  from the baseline to the chosen item, a positive value shows in
  increase in correlation, a negative value implies a decrease in
  correlation. The stars represent the statistical significance level
  of each expected change. The footer of the table shows the goodness
  of fit statistics for the analytical model.} \label{table:kcoefficients}
\end{center}
\end{table*}

Notice the
high $R^2$ value and small RMSE values for each model, showing that
the deviation of the data to this analytical model is small. This implies that the
analytical model is appropriate for the correlation data analysis.

The table shows each variable (Expected Accuracy Function, Data, Classifier)
using the baseline values High Prob for Function, Emotions for Data and
ECC for Classifier. The entries in the table show the bias (change in mean
variance stabilised Kendall's tau correlation) when changing the
variable from the baseline to the given value. A positive value implies that the
correlation has increased, implying that the candidate function and accuracy
are closer to being weakly monotonic, while a negative value implies a
reduction in correlation.

The
table shows that TG reduces the correlation (shown by its
negative coefficient), and that Shannon Entropy and Collision Entropy
increase the correlation (shown by their positive coefficients) for
each of the Exact, Hamming and Jaccard similarities. The stars
associated to each coefficient show the range of the p-value for the
hypothesis test, testing if there is evidence that the coefficient is
not zero.  No significance implies that there is no evidence and hence
using the acceptance method is no different to the baseline, and
significance implies a difference to the baseline HP. 

The results imply that we should be using HP for
Exact, Shannon Entropy for Jaccard and Collision Entropy for Hamming
similarity.

The Data section shows that changing the data set does
have an effect on the expected correlation, with Enron having the
greatest effect. 

The Classifier section indicates that using an Independent
classifier leads to significantly worse correlation at the 1\% level
for Exact, Hamming and Jaccard. CT also provides a drop in expected
correlation, but is only significant at the 1\% level when using
Hamming similarity. Since we generally 
expect that the order from lowest to highest (due to the complexity of
each method) would be Independent, CT, ECC, the analysis shows that a more accurate classifier leads to higher
expected correlation.

Note that since the analytical model is modeling each variable
independently, the significant results for Data and Classifier do not impact the
significant results for Expected Accuracy functions.

\stepcounter{AnalysisNumber}
\subsection{Analysis \theAnalysisNumber: Robustness Analysis for Relative Accuracy}

\label{sec:krobust}

Robustness is very important when measuring expected accuracy, since we require
that the results have only small variation based on minor changes in
experimental parameters (i.e., the correlation between the accuracy
and candidate expected accuracy should not be effected by a change in data,
method, or evaluation function, etc). In other words, we want these
different parameters to have negligible effect on the correlation
between the classifier accuracy and the estimated expected accuracy.

For this analysis, we examined each candidate function independently,
and examined the effect of several data features, change in
classifier, and change in evaluation function on the candidate expected accuracy to
accuracy.
The
results of the analysis are in Table \ref{table:krobust}.

\begin{table*}
\begin{center}
\begin{tabular}{l D{)}{)}{13)1} D{)}{)}{13)1} D{)}{)}{13)0} D{)}{)}{13)1} }
\toprule
 & \multicolumn{1}{c}{HP} & \multicolumn{1}{c}{TG} & \multicolumn{1}{c}{SE} & \multicolumn{1}{c}{CE} \\
\midrule
\multicolumn{4}{l}{\textit{Data Features Expected Gradient (Zero gradient means no effect)}} \\
\zspace{}Label Count            & -0.046 \; (0.023)^{*}   & -0.078 \; (0.022)^{***} & -0.024 \; (0.019) & -0.038 \; (0.022)^{*} \\
\zspace{}Label Comb             & 0.011 \; (0.007)^{*}    & 0.022 \; (0.006)^{***}  & 0.005 \; (0.005)  & 0.009 \; (0.006)      \\
\zspace{}Label Card             & -0.471 \; (0.292)       & -0.959 \; (0.282)^{***} & -0.208 \; (0.237) & -0.374 \; (0.277)     \\
\zspace{}Feature Count          & -0.001 \; (0.001)^{*}   & -0.002 \; (0.001)^{***} & -0.000 \; (0.000) & -0.001 \; (0.001)     \\[0.5em]
\multicolumn{4}{l}{\textit{Classifier Bias (Baseline: ECC)}} \\
\zspace{}Independent & -0.092 \; (0.034)^{***} & -0.112 \; (0.032)^{***} & -0.039 \; (0.027) & -0.061 \; (0.032)^{*} \\
\zspace{}CT          & -0.040 \; (0.034)       & -0.077 \; (0.032)^{**}  & 0.015 \; (0.027)  & -0.023 \; (0.032)     \\[0.5em]
\multicolumn{4}{l}{\textit{Similarity Function Bias (Baseline: Exact)}} \\
\zspace{}Hamming            & -0.009 \; (0.034)       & -0.023 \; (0.032)       & -0.000 \; (0.027) & 0.001 \; (0.032)      \\
\zspace{}Jaccard            & -0.077 \; (0.034)^{**}  & -0.088 \; (0.032)^{***} & -0.035 \; (0.027) & -0.059 \; (0.032)^{*} \\
\midrule
R$^2$                  & 0.436                   & 0.478                   & 0.224             & 0.311                 \\
Adj. R$^2$             & 0.336                   & 0.385                   & 0.086             & 0.188                 \\
Num. obs.              & 54                      & 54                      & 54                & 54                    \\
RMSE                   & 0.101                   & 0.097                   & 0.082             & 0.096                 \\
\bottomrule
\multicolumn{5}{l}{\scriptsize{$^{***}p<0.01$, $^{**}p<0.05$, $^*p<0.1$}}
\end{tabular}
\caption{The expected change in variance stabilised Kendall's $\tau$
  correlation when adjusting four data features (Label Count, Label Comb,
  Label Card, Feature Count), or changing Classifier or Similarity Function
  (standard error in parentheses), independent of the other variable values. A value of 0
  represents no change in the correlation when changing the Data Feature, or when changing from the
  baseline to the chosen item, a positive value shows in increase in
  correlation, a negative value implies a decrease in correlation. The
  stars represent the statistical significance level of each expected
  change. The footer of the table shows the goodness of fit statistics
  for the analytical
  model.}
\label{table:krobust}
\end{center}
\end{table*}

The section \textit{Data Features Expected Gradient}
within Table \ref{table:krobust}
contains Label
Count (the number of labels in the multi-label data), Label Comb (the
number of unique label combinations found in the data), Label Card
(the average label cardinality for the data set), and Feature Count
(the number of prediction features in the data). Each of these are
either positive real or positive integer variables, therefore the
estimated analytical model coefficient for each is its gradient (the expected increase
in Fisher transformed correlation when the associated variable
increases by 1).  For example, note that the coefficient for Label
Count in the HP column is $-0.046$, telling us that we expect the
Fisher transformed correlation for HP to decrease by $0.046$ with each
additional label in a dataset, implying that there is a dependence of
the correlation on the the number of labels in the data.

Note that we are not concerned about the baseline being non-zero; to
show robustness, we are required to examine if changes to the
experiment variables changes the correlation therefore we are only
concerned with the coefficients of the remaining additive effects. An
ideal expected accuracy function will have zero for all coefficients
(implying that any change in experiment variables does not effect its
correlation to accuracy).

We find that the least robust is TG, having statistically significant
non-zero coefficients for all but the Hamming coefficient (meaning
that if we change the data, classifier, or similarity function, the
acceptance score will correlate differently to the prediction
accuracy). The most robust is SE, having no
statistically significant non-zero coefficients, followed closely by
CE, having coefficients with significance only at the 10\% level.  

It is interesting to see that HP is affected by changing the
classifier from ECC to Independent, or changing the accuracy metric
to Jaccard. Also the sample gradients for Label Count, Label
Card and Feature count are negative (as these increase
in the data, the correlation will reduce), while the sample gradient
for Label comb is positive (as the number of label
combinations increase, the correlation will increase) for all
acceptance parameters. However, note that many of these results are not
statistically significant.

The goodness of fit statistics (shown at the bottom of
the table) are all poor, especially for SE and CE. This is a good
sign, showing that the correlation has little association with the
data parameters, classifier and accuracy metric. We conclude this
section noting that SE is the most robust in terms of Kendall's
correlation to accuracy.

\section{Absolute Expected Accuracy}
\label{sec:absolute}

We now examine the potential for each candidate expected accuracy function to
provide \emph{absolute} accuracy, where the value of the
candidate expected accuracy is an estimation of the accuracy of the prediction.
An accurate absolute accuracy score provides us with a good estimate of the
accuracy of a prediction.
To measure the accuracy of each candidate for absolute accuracy,
we measure the Pearson correlation between the candidate expected accuracy
estimate and the accuracy of the prediction. A high correlation
implies that there is a linear relationship between the prediction
accuracy and the candidate expected accuracy.

\stepcounter{AnalysisNumber}
\subsection{Analysis \theAnalysisNumber: Marginal Correlation Analysis for Absolute Accuracy}

We begin our evaluation of each candidate expected accuracy function
with a marginal analysis of each of the experiment variables. This
analysis is similar to that presented on Table
\ref{table:kcoefficients}, but instead we are examining the response
of the Pearson correlation between prediction accuracy and acceptance
function. A high Pearson correlation implies that there is a linear
relationship between the candidate score and the accuracy of a given
prediction, which in turn tells us that we are able to use the
candidate to estimate the expected accuracy of the
prediction. Therefore, we want to find which of the candidates leads
to the greatest Pearson correlation.

In the previous analysis we left out the Minimum Entropy (ME) and
Chi-Squared (CS) candidates since they provided the same ordering of
values to HP and CE respectively but absolute results may be different, hence we include them in this section.
Results are presented in Table \ref{table:pcoefficients}.

\begin{table*}
\begin{center}
\begin{tabular}{l D{)}{)}{13)3} D{)}{)}{13)3} D{)}{)}{13)3} }
\toprule
 & \multicolumn{1}{c}{Exact} & \multicolumn{1}{c}{Jaccard} & \multicolumn{1}{c}{Hamming} \\
\midrule
\multicolumn{4}{l}{\textit{Expected Accuracy Function Bias (Baseline: High Prob)}} \\
\zspace{}TG              & -0.073 \; (0.026)^{\star\star\star} & -0.093 \; (0.023)^{\star\star\star} & -0.096 \; (0.017)^{\star\star\star} \\
\zspace{}SE              & 0.000 \; (0.026)        & 0.067 \; (0.023)^{\star\star\star}  & 0.025 \; (0.017)        \\
\zspace{}CE              & -0.006 \; (0.026)       & 0.029 \; (0.023)        & 0.024 \; (0.017)        \\
\zspace{}ME              & -0.024 \; (0.026)       & -0.017 \; (0.023)       & -0.010 \; (0.017)       \\
\zspace{}CS              & 0.011 \; (0.026)        & 0.037 \; (0.023)        & 0.024 \; (0.017)        \\[0.5em]
\multicolumn{4}{l}{\textit{Data Bias (Baseline: Emotions)}} \\
\zspace{}Scene             & 0.117 \; (0.026)^{\star\star\star}  & 0.054 \; (0.023)^{\star\star}   & 0.113 \; (0.017)^{\star\star\star}  \\
\zspace{}Yeast             & 0.214 \; (0.026)^{\star\star\star}  & 0.102 \; (0.023)^{\star\star\star}  & 0.126 \; (0.017)^{\star\star\star}  \\
\zspace{}Stare             & -0.028 \; (0.026)       & -0.043 \; (0.023)^{\star}   & -0.004 \; (0.017)       \\
\zspace{}Enron             & 0.073 \; (0.026)^{\star\star\star}  & -0.464 \; (0.023)^{\star\star\star} & 0.136 \; (0.017)^{\star\star\star}  \\
\zspace{}Slashdot          & -0.032 \; (0.026)       & -0.113 \; (0.023)^{\star\star\star} & -0.020 \; (0.017)       \\[0.5em]
\multicolumn{4}{l}{\textit{Classifier Bias (Baseline: ECC)}} \\  
\zspace{}Independent & -0.071 \; (0.019)^{\star\star\star} & -0.089 \; (0.016)^{\star\star\star} & -0.087 \; (0.012)^{\star\star\star} \\
\zspace{}CT          & -0.019 \; (0.019)       & -0.025 \; (0.016)       & -0.056 \; (0.012)^{\star\star\star} \\
\midrule
R$^2$                  & 0.634                   & 0.900                   & 0.771                   \\
Adj. R$^2$             & 0.587                   & 0.888                   & 0.742                   \\
Num. obs.              & 108                     & 108                     & 108                     \\
RMSE                   & 0.079                   & 0.070                   & 0.051                   \\
\bottomrule
\multicolumn{4}{l}{\scriptsize{$^{\star\star\star}p<0.01$, $^{\star\star}p<0.05$, $^\star p<0.1$}}
\end{tabular}
\caption{The expected change in variance stabilised Pearson's
  correlation for the three variables (Expected Accuracy function, Data,
  Classifier) and standard error in parentheses, with respect to each
  variable's baseline, independent of the other variable values. A
  value of 0 represents no change in the correlation when changing
  from the baseline to the chosen item, a positive value shows in
  increase in correlation, a negative value implies a decrease in
  correlation. The stars represent the statistical significance level
  of each expected change. The footer of the table shows the goodness
  of fit statistics for the analytical model.} \label{table:pcoefficients}
\end{center}
\end{table*}

All candidate expected accuracy comparisons are relative to HP. Each
Expected Accuracy Bias, Data Bias and Classifier Bias is the change in
mean Fisher transformed Pearson correlation between the predicted
expected accuracy and
true expected accuracy; a positive bias means that the change of variable
has lead to an increase in correlation, meaning that the confidence function
is able to provide a better estimates of expected accuracy.

The \textit{Expected Accuracy} portion of the table shows that TG is highly
significant for Exact, Hamming and Jaccard Similarity and the
coefficients are negative; thus, TG is significantly
worse that HP. The only other statistically significant coefficient is
for SE when using Jaccard Similarity (positive). 

In the \textit{Data} section, we find that Scene,
Yeast and Enron all have a significant effect when using each of the Exact,
Hamming and Jaccard similarities, and that Stare and Slashdot also
have a significant effect when using Jaccard Similarity.

In the \textit{Classifier} portion of the table, all of the
coefficients are negative, implying that changing from ECC to either
Independent or CT will reduce the correlation between accuracy and
acceptance, where Independent leads to a greater statistically significant drop.

Finally the goodness of fit statistics show that the three variables
(Expected Accuracy function, Data and Classifier) explain the variability in
correlation very well when using Jaccard Similarity (shown by the high
$R^2$ value), followed by Hamming and Exact. The $R^2$ value for Exact
is on the low side, indicating that there may additional variables or
interactions that we have not taken into account.  The small RMSE
values give an indication of the expected analytical model error,
where all are similar. This leads us to believe that the low $R^2$
value for Exact is due to the accuracy-candidate expected accuracy correlation having
greater total variance. 

Thus, results indicate that SE should be used when measuring acceptance for Jaccard Similarity. There was no
significant evidence for using other acceptance parameters over the baseline HP.

\stepcounter{AnalysisNumber}
\subsection{Analysis \theAnalysisNumber: Robustness Analysis for Absolute Accuracy}

We now investigate the robustness of each candidate function when used for absolute accuracy.
The results of the analysis are shown in Table
\ref{table:probust}; interpretation is similar to \Tab{table:krobust} of \Sec{sec:krobust}: split into five sections, the first four
containing the marginal effects and the last section containing
goodness of fit statistics. Each estimated coefficient is accompanied
with a standard error directly below it in parentheses.

\begin{table}
\begin{center}
\begin{tabular}{l D{.}{.}{2.5} D{.}{.}{2.6} D{.}{.}{2.4} D{.}{.}{2.4} D{.}{.}{2.5} D{.}{.}{2.4} }
\toprule
 & \multicolumn{1}{c}{HP} & \multicolumn{1}{c}{TG} & \multicolumn{1}{c}{SE} & \multicolumn{1}{c}{CE} & \multicolumn{1}{c}{ME} & \multicolumn{1}{c}{CS} \\
\midrule
\multicolumn{7}{l}{\textit{Data Features Expected Gradient (Zero gradient means no effect)}} \\
\zspace{}Label Count            & -0.020      & -0.079^{**}  & 0.029     & 0.009   & -0.039      & 0.028   \\
                       & (0.031)     & (0.030)      & (0.023)   & (0.029) & (0.030)     & (0.029) \\
\zspace{}Label Comb             & 0.003       & 0.021^{**}   & -0.010    & -0.006  & 0.008       & -0.011  \\
                       & (0.009)     & (0.009)      & (0.007)   & (0.008) & (0.008)     & (0.008) \\
\zspace{}Label Card             & -0.068      & -0.902^{**}  & 0.506^{*} & 0.302   & -0.295      & 0.525   \\
                       & (0.388)     & (0.381)      & (0.295)   & (0.364) & (0.376)     & (0.365) \\
\zspace{}Feature Count          & -0.000      & -0.002^{**}  & 0.001^{*} & 0.001   & -0.001      & 0.001   \\
                       & (0.001)     & (0.001)      & (0.001)   & (0.001) & (0.001)     & (0.001) \\[0.5em]
\multicolumn{4}{l}{\textit{Classifier Bias (Baseline: ECC)}} \\
\zspace{}Independent & -0.100^{**} & -0.134^{***} & -0.032    & -0.068  & -0.104^{**} & -0.055  \\
                       & (0.045)     & (0.044)      & (0.034)   & (0.042) & (0.043)     & (0.042) \\
\zspace{}CT          & -0.046      & -0.090^{**}  & 0.026     & -0.026  & -0.053      & -0.010  \\
                       & (0.045)     & (0.044)      & (0.034)   & (0.042) & (0.043)     & (0.042) \\[0.5em]
\multicolumn{4}{l}{\textit{Similarity Function Bias (Baseline: Exact)}} \\
\zspace{}Hamming            & -0.025      & -0.048       & -0.000    & 0.005   & -0.010      & -0.011  \\
                       & (0.045)     & (0.044)      & (0.034)   & (0.042) & (0.043)     & (0.042) \\
\zspace{}Jaccard            & -0.078^{*}  & -0.098^{**}  & -0.012    & -0.044  & -0.072      & -0.052  \\
                       & (0.045)     & (0.044)      & (0.034)   & (0.042) & (0.043)     & (0.042) \\
\midrule
R$^2$                  & 0.424       & 0.459        & 0.261     & 0.341   & 0.453       & 0.327   \\
Adj. R$^2$             & 0.322       & 0.363        & 0.129     & 0.224   & 0.355       & 0.207   \\
Num. obs.              & 54          & 54           & 54        & 54      & 54          & 54      \\
RMSE                   & 0.134       & 0.131        & 0.102     & 0.126   & 0.130       & 0.126   \\
\bottomrule
\multicolumn{7}{l}{\scriptsize{$^{***}p<0.01$, $^{**}p<0.05$, $^*p<0.1$}}
\end{tabular}
\caption{The expected change in variance stabilised Pearson's
  correlation when adjusting four data features (Label Count, Label Comb,
  Label Card, Feature Count), or changing Classifier or Similarity Function
  (standard error in parentheses), independent of the other variable values. A value of 0
  represents no change in the correlation when changing the Data Feature, or when changing from the
  baseline to the chosen item, a positive value shows in increase in
  correlation, a negative value implies a decrease in correlation. The
  stars represent the statistical significance level of each expected
  change. The footer of the table shows the goodness of fit statistics
  for the analytical
  model.}
\label{table:probust}
\end{center}
\end{table}

TG is affected by most of the variables (being significant at the 5\% level for most
coefficients). Change in any of the variable values for both CE and CS has no
significant effect on the accuracy-confidence correlation.  Both HP
and ME show a significant effect for changing the classifier from ECC
to Independent and HP has the addition significant effect when
changing from Exact to Jaccard Similarity. SE shows to have a
significant effect on two of the data variables (Label card and
Feature count), but it is interesting to see that this effect is
positive, meaning that the correlation between the candidate function and accuracy
when using SE increases as the Label cardinality or Feature count
increases. 

Examining the Classifier variable, we find that
these results reinforce our belief that the accuracy-candidate correlation is dependent on the complexity of the classifier.

Surprisingly, SE has the lowest $R^2$. therefore the accumulation of all variables must explain
more of the variance for CE and CS when compared to SE, even though SE
has two significant coefficients.

Hence we see that CE and CS are the most robust in this context.  SE was
affected by some of the data parameters, but it showed that the
correlation improves as the Label cardinality increases, which is
usually a function of data size. Therefore, SE should be better suited
to large data sets than small data sets.

\stepcounter{AnalysisNumber}
\subsection{Analysis \theAnalysisNumber: Mixture Analysis for Absolute Accuracy}

So far we have examined the potential for each of the candidate
expected accuracy function to be used to measure absolute acceptance (we examined correlation, showing a linear relationship), but
we have not examined how to map the candidate functions to an
estimated accuracy.
In this section, we will train models to estimate the Exact, Hamming
and Jaccard accuracy using the candidate functions, to assess how
well each is able to provide estimates of absolute accuracy.

To estimate the expected accuracy, we build a logistic regression model. Both
Exact Match and Hamming similarity lead to this because they can be
modelled as Bernoulli and Binomial random variables, respectively.

The features of the model are the predicted expected accuracy provided by the
candidate function (continuous),
the data set name (categorical), and the classification method
(categorical). Since the accuracy results vary heavily based on the
data set, we also included an interaction term between the candidate function
and data set, giving the expected accuracy model:
\begin{align}
\label{eq:model}
  \text{logit}(p_{ijx}) = \beta_{i}\text{candidate}(x) +  \text{data}_i + \text{classifier}_j
\end{align}
where $\text{candidate}(x)$ is the value provided by the candidate function for instance
$x$, $\beta_{i}$ is the candidate coefficient for data set $i$,
$\text{data}_i$ is a bias term for data set $i$ ($i$ is either
Emotions, Scene, Yeast, Stare, Enron, or Slashdot),
$\text{classifier}_j$ is the bias term for classifier $j$ ($j$ is
either Independent, CT or ECC), $\text{logit}$ is the logit function,
and $p_{ijx}$ is the estimated accuracy of the prediction for instance $x$.

Both Hamming Similarity and Exact Match are suited to logistic regression,
since they are both proportions (Exact is either 1 or 0, and Hamming
is represented as the proportion of correct labels).
Modelling Jaccard Similarity is not as straightforward. Since both the numerator and denominator (see \Eq{eq:jaccard}) 
are affected by the prediction, we can not simply model the
similarity as a proportion. We can instead decompose
the Jaccard Similarity into the number of true and false positives and
negatives:
\begin{align}
  S_{\text{Jaccard}}(y, \hat{y}) = \frac{\text{TP}}{\text{TP} + \text{FP} + \text{FN}}
\end{align}
where TP is the number of labels that were predicted as
positive and are actually positive, $\text{FP}$ is the number of
labels that were predicted as positive, but are actually negative,
$\text{FN}$ is the number of labels that are predicted as negative,
but are actually positive, thus
$\text{TN} = N - \text{TP} - \text{FN} - \text{FP}$ is the number of
labels that were predicted as negative and are actually negative ($N$ is the number of labels).  The
number of labels $N$ is constant for each prediction, therefore the
set $\{\text{TP}, \text{TN}, \text{FP}, \text{FN}\}$ is equivalent to a
multinomial distribution with $N$ trials. Modelling this multinomial distribution will provide estimates for the four true/false positive/negative rates $p_{\text{TP}}$, $p_{\text{TN}}$, $p_{\text{FP}}$, $p_{\text{FN}}$ allowing us to model the Jaccard Similarity finally as
\begin{align}
  S_{\text{Jaccard}}(\y, \ypred) = \frac{p_{\text{TP}}}{p_{\text{TP}} + p_{\text{FP}} + p_{\text{FN}}}
\end{align}
which can be modelled with as a multinomial distribution using a multinomial
logistic regression with the same covariates as in equation
\ref{eq:model}. The candidate function scores can then be used to predict
the true/false positive/negative proportions and then reconstruct the
estimate of the Jaccard Similarity.

Using the predictive models for Exact, Hamming and Jaccard Similarity,
we examine the utility of HP, SE and CE candidate functions
(which so far showed merit) to
predict the accuracy. We will also examine the accuracy of a
combined model using the model in equation \ref{eq:model} where all
candidate functions are linearly combined, rather than only one.

The complete data set (containing all multi-label data sets) was
randomly split into 50\% training and 50\% testing sets and 10-fold
cross validation was used on the training set to determine the
ridge regularisation parameters of each model.  The complete randomised
training/testing process was replicated 20 times to obtain
distributions of expected accuracy for each fitted model.

To evaluate the results, we must remember that each predicted value
is an expected accuracy value; we can evaluate the prediction by
comparing it to the actual accuracy for the given instance. For our
evaluation, we are also interested in the value of the actual
accuracy (e.g. there will be cases when we want a more accurate model for
only the high accuracy results rather then all results). So we provide
measurements for various ranges of accuracy.
The estimated expected accuracies were divided into 0.1 intervals and
the associated instances in each interval were used to compute the
true expected accuracy (the intervals were size 0.05 for Hamming
confidence due to the smaller range of scores). This was applied to
the 20 replicates to obtain a distribution over the expected
accuracy. The 95\% interval of the expected accuracy distribution is
reported for each of the Hamming, Jaccard and Exact Match confidence
score intervals in Tables \ref{tab:hammci}, \ref{tab:jacmci} and
\ref{tab:emmci} respectively.  Ideally, the 95\% expected accuracy
interval should be contained within the given estimated expected
accuracy interval.

First examining the Hamming table (Table \ref{tab:hammci}), the
expected Hamming accuracy stays within the estimated expected accuracies
for all ranges except for $(0.7,0.75]$ for High Prob, $(0.65,0.7]$ for
Shannon Entropy, $(0.65,0.7]$, $(0.8,0.85]$ for Collision Entropy, and
$(0.65,0.7]$ for the mixture of each candidate function.
Each of the four methods of predicting the expected accuracy have
performed well. Collision Entropy shows to have the most interval
matches (shown in bold), but it provides an overestimate in the
range of $(0.8,0.85]$ which is not desired. Shannon Entropy provides
the next most interval matches.

Next examining the Jaccard accuracy interval table (Table \ref{tab:jacmci}),
the expected Jaccard accuracy lies within the computed accuracy
bounds for $(0.4, 0.5]$ and $(0.2, 0.3]$ using HP, SE and CE and $(0.9,1]$
for the mixture of confidence functions. Each model does provide a
greater true expected Jaccard accuracy for each computed confidence
range, except for HP, SE and CE for the interval $(0.9, 1]$ and all
models for the interval $(0,0.1]$. The mixture is preferred in this
case, due to it providing la large number of interval matches, and not
overestimating the high accuracy values.

Finally, the Exact Match interval table (Table \ref{tab:emmci})
provides the expected Exact Match for each of the models'
computed intervals.
The correct predicted intervals were provided at
$(0.2,0.3]$ for HP, $(0.1, 0.2]$ for SE,
$(0.2,0.3]$ and $(0.1, 0.2]$ for CE, and
$(0.4,0.5]$, $(0.3,0.4]$, $(0.1, 0.2]$ and $(0,0.1]$ for the mixture.
Each expected accuracy interval was within the range of the computed
interval, or greater than the interval for all but three cases for HP,
two cases for SE, three cases for CE and two cases for the
mixture. For the two cases where the mixture overestimates, the error
is small (for the predicted expected accuracy interval $(0.9, 1.0]$ the
expected Exact Match accuracy interval is $(0.887, 0.940]$, and for the
predicted expected accuracy of $(0.7, 0.8]$ the expected accuracy interval is
$(0.691, 0.806]$). Again, we prefer the Mixture, due to the most
interval matches and the fewest overestimates.

\newcommand{\bb}[1]{\textbf{#1}}
\newcommand{\uu}[1]{\underline{#1}}

\begin{table}[ht]
\centering
\caption{The 95\% interval for the expected Hamming accuracy for the
  given range of predicted Hamming confidence when modeled on each of
  High Prob (HP), Shannon Entropy (SE), Collision Entropy (CE) and a
  mixture of all confidence functions (MIX). The bold expected Hamming
  intervals are the closest match to the predicted intervals. The
  underlined intervals have a lower bound that is lower than the
  predicted interval.}

\begin{tabularx}{0.9\linewidth}{rZZZZ}
  \toprule
  Exp Acc & HP & SE & CE & MIX \\ 
  \midrule
  (0.65,0.7] &       NA     & 0.688, 0.746 & \bb{0.657, 0.754} & \uu{0.641, 0.737} \\ 
  (0.7,0.75] & 0.720, 0.772 & \bb{0.707, 0.742} & 0.720, 0.747 & 0.709, 0.733 \\ 
  (0.75,0.8] & 0.750, 0.764 & 0.767, 0.782 & \bb{0.765, 0.789} & 0.763, 0.780 \\ 
  (0.8,0.85] & 0.808, 0.830 & \bb{0.815, 0.848} & \uu{0.792, 0.824} & 0.822, 0.837 \\ 
  (0.85,0.9] & 0.873, 0.885 & \bb{0.866, 0.884} & \bb{0.864, 0.882} & 0.871, 0.888 \\ 
  (0.9,0.95] & 0.932, 0.940 & 0.923, 0.938 & \bb{0.925, 0.945} & 0.929, 0.935 \\ 
  (0.95,1]   & 0.959, 0.962 & 0.955, 0.959 & \bb{0.957, 0.965} & 0.961, 0.964 \\ 
   \bottomrule
\end{tabularx}
\label{tab:hammci}
\end{table}

\begin{table}[ht]
\centering
\caption{The 95\% interval for the expected Jaccard accuracy for the given range of
  predicted Hamming confidence when modeled on each of High Prob (HP),
  Shannon Entropy (SE), Collision Entropy (CE) and a mixture of all
  confidence functions (MIX). The bold expected Jaccard
  intervals are the closest match to the predicted intervals. The
  underlined intervals have a lower bound that is lower than the
  predicted interval.}
\begin{tabularx}{0.9\linewidth}{rZZZZ}
    \toprule
  Exp Acc & HP & SE & CE & MIX \\   
  \midrule
  (0,0.1]   & 0.362, 0.399 & 0.362, 0.398 & 0.362, 0.399 & 0.328, 0.374 \\ 
  (0.1,0.2] & 0.185, 0.363 & 0.196, 0.301 & 0.183, 0.329 & \uu{\bb{0.092, 0.136}} \\ 
  (0.2,0.3] & 0.236, 0.257 & 0.226, 0.263 & 0.246, 0.280 & \bb{0.259, 0.308} \\ 
  (0.3,0.4] & 0.384, 0.480 & 0.431, 0.483 & 0.383, 0.462 & \bb{0.385, 0.429} \\ 
  (0.4,0.5] & \bb{0.450, 0.495} & 0.449, 0.488 & 0.439, 0.483 & 0.474, 0.526 \\ 
  (0.5,0.6] & \bb{0.579, 0.639} & 0.596, 0.679 & 0.584, 0.642 & 0.592, 0.647 \\ 
  (0.6,0.7] & \bb{0.677, 0.771} & 0.779, 0.861 & 0.726, 0.872 & 0.698, 0.772 \\ 
  (0.7,0.8] & 0.880, 0.953 & 0.894, 0.956 & 0.909, 0.960 & \bb{0.845, 0.895} \\ 
  (0.8,0.9] & 0.904, 0.968 & 0.889, 0.949 & \bb{0.858, 0.960} & 0.924, 0.973 \\ 
  (0.9,1]   & \uu{0.785, 1.000} & \uu{0.817, 0.952} & \uu{0.771, 1.000} & \bb{0.913, 0.968} \\ 
   \bottomrule
\end{tabularx}
\label{tab:jacmci}
\end{table}

\begin{table}[ht]
\caption{The 95\% interval for the expected Exact Match accuracy for the given range of
  predicted Hamming confidence when modeled on each of High Prob (HP),
  Shannon Entropy (SE), Collision Entropy (CE) and a mixture of all
  confidence functions (MIX). The bold expected Exact Match
  intervals are the closest match to the predicted intervals. The
  underlined intervals have a lower bound that is lower than the
  predicted interval.}
\centering
\begin{tabularx}{0.9\linewidth}{rZZZZ}
  \toprule
  Exp Acc & HP & SE & CE & MIX \\ 
  \midrule
  (0,0.1]   & \bb{0.051, 0.105} & 0.066, 0.108 & 0.066, 0.106 & 0.056, 0.094 \\ 
  (0.1,0.2] & 0.171, 0.201 & \bb{0.126, 0.174} & 0.134, 0.179 & 0.108, 0.145 \\ 
  (0.2,0.3] & 0.205, 0.236 & \uu{0.199, 0.222} & 0.203, 0.231 & \bb{0.254, 0.305} \\ 
  (0.3,0.4] & \uu{0.256, 0.314} & \bb{0.317, 0.401} & 0.334, 0.412 & 0.319, 0.353 \\ 
  (0.4,0.5] & 0.491, 0.598 & 0.479, 0.528 & 0.451, 0.555 & \bb{0.402, 0.474} \\ 
  (0.5,0.6] & 0.555, 0.659 & 0.541, 0.638 & \uu{\bb{0.490, 0.583}} & 0.560, 0.648 \\ 
  (0.6,0.7] & \uu{0.579, 0.670} & 0.652, 0.750 & \uu{0.568, 0.700} & \bb{0.617, 0.710} \\ 
  (0.7,0.8] & 0.724, 0.839 & 0.786, 0.880 & 0.728, 0.815 & \uu{\bb{0.691, 0.806}} \\ 
  (0.8,0.9] & 0.893, 0.955 & 0.842, 0.931 & 0.873, 0.924 & \bb{0.835, 0.924} \\ 
  (0.9,1]   & \uu{0.860, 0.956} & \uu{0.820, 0.894} & \uu{0.854, 0.964} & \uu{\bb{0.887, 0.940}} \\ 
   \bottomrule
\end{tabularx}
\label{tab:emmci}
\end{table}

\begin{figure*}
\includegraphics[scale=0.415]{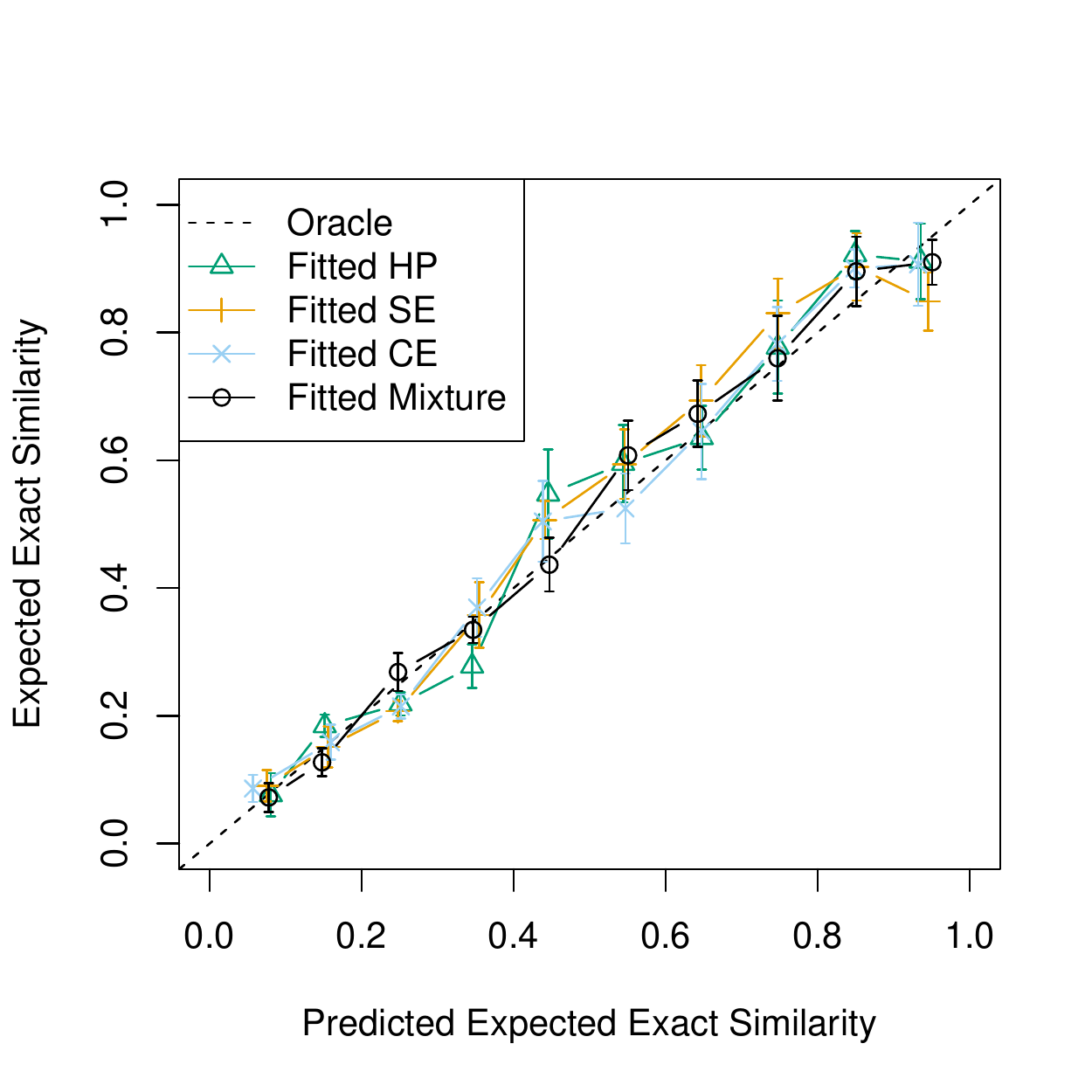}
\includegraphics[scale=0.415]{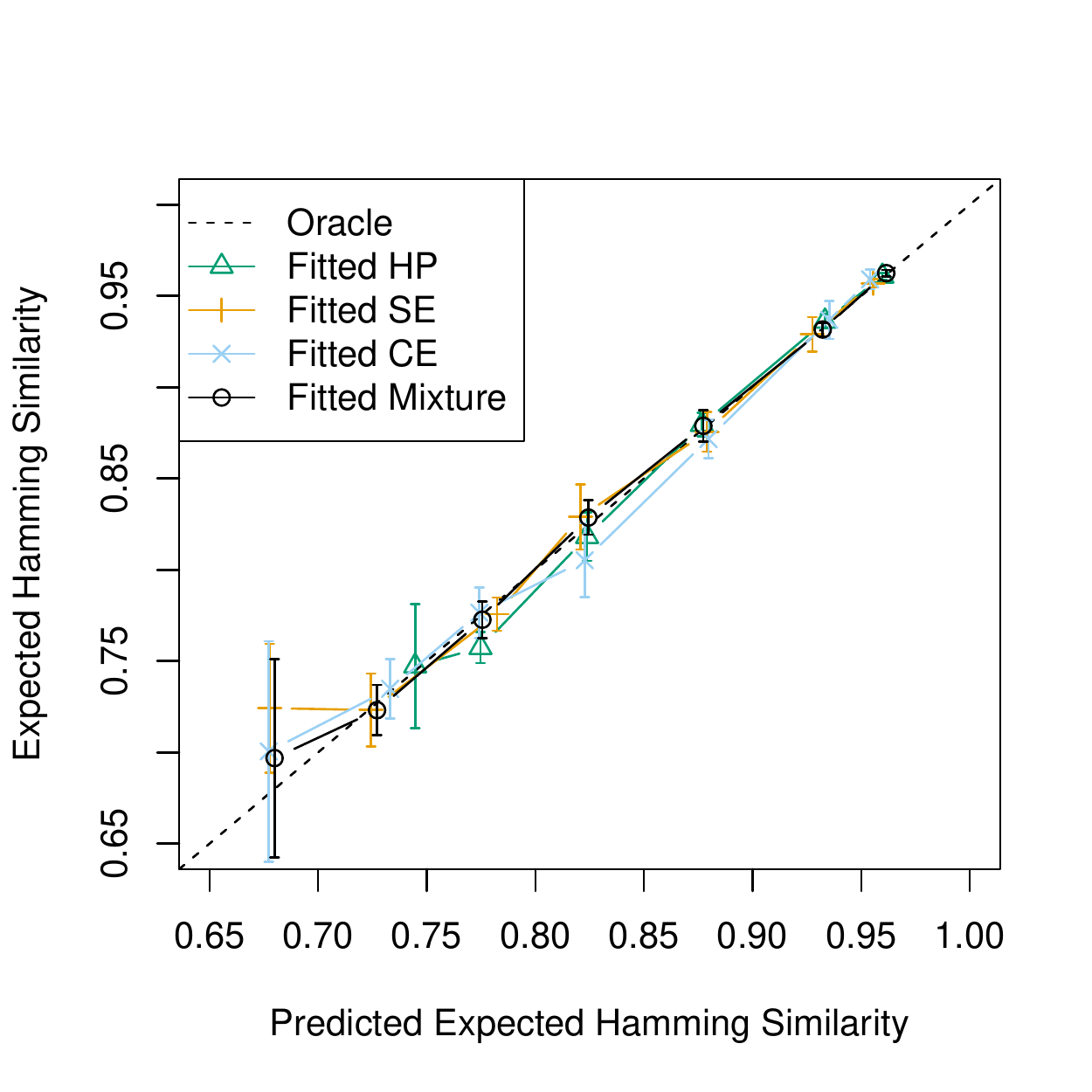}
\includegraphics[scale=0.415]{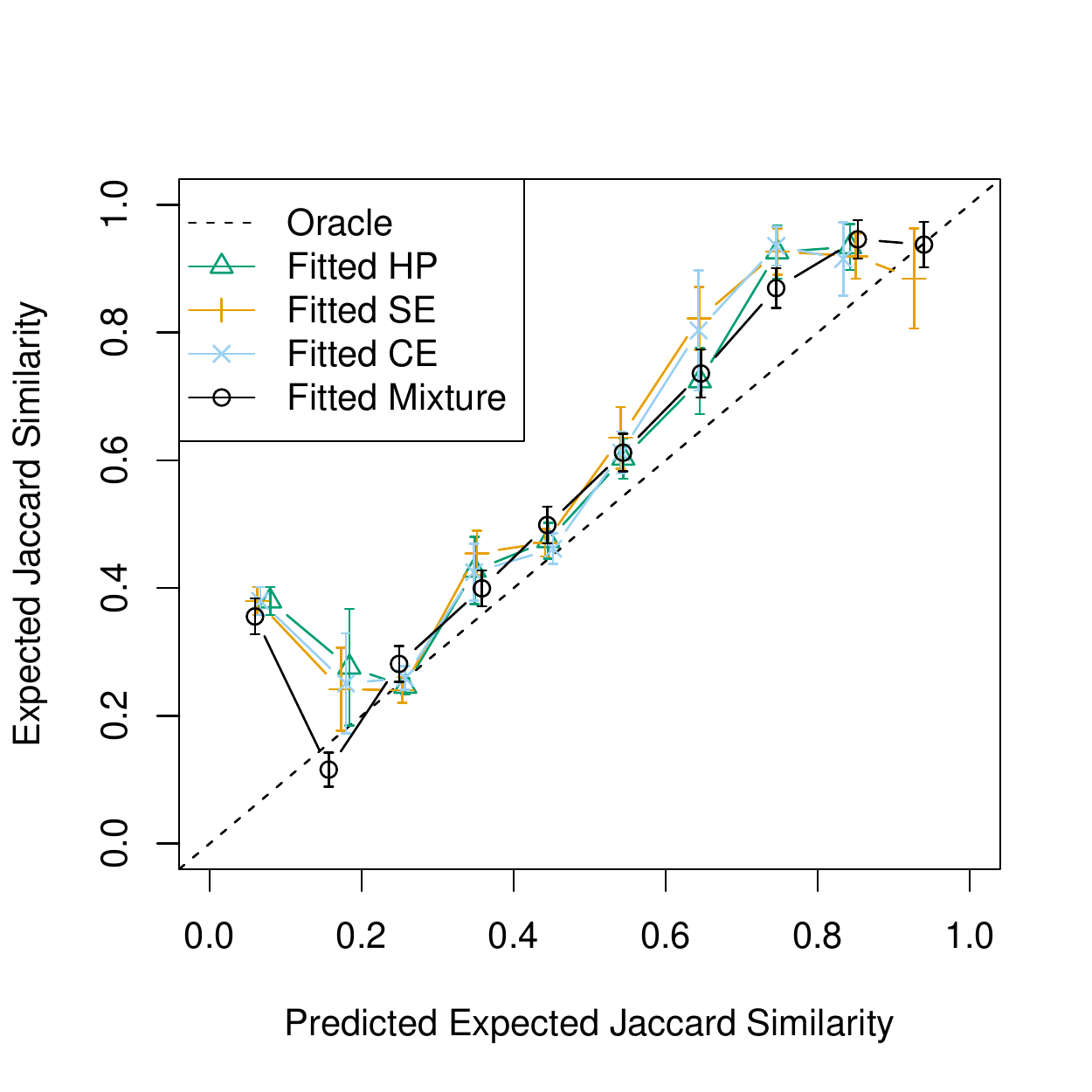} \\
\caption{The expected accuracy versus the predicted expected accuracy
  for the Exact, Hamming and Jaccard Similarity functions, using
  the candidate functions HP, SE, CE, or a mixture of all. The
  error bars show the 95\% confidence interval for the expected
  accuracy. The Oracle line shows the ideal values.}
\label{fig:scoreacc}
\end{figure*}

Figure \ref{fig:scoreacc} presents the expected accuracy for a given
predicted expected accuracy score. The ideal expected accuracy function will provide a
straight line travelling along the diagonal, meaning that the
estimated expected accuracy has a one to one mapping with the true
expected accuracy.

We see that the Mixture curve is always the closest to the diagonal line, or has small error, where the other
functions may be close to the diagonal at some points, but have large
deviations at other points. Both Exact and Hamming similarities have
close fits to the diagonal, but Jaccard has a large error at the lower
end.It is unclear why
this has occurred, but being at the lower end means that it will not
impact the use of the expected accuracy functions.

To summarise the results, we have found that when fitting the expected
accuracy function (consisting of the candidate functions) to data, any
of the candidate functions will provide a good model for Hamming
Similarity, and a mixture of the set of candidate functions provides
the best estimates for Jaccard and Exact Match.

\section{Conclusion}
\label{sec:conclusion}

Probabilistic multi-label classification provides us with a
distribution over the power set of labels. The probability of the
mode labelset is commonly used as an estimate of the prediction
confidence, however this measurement is not directly dependent on the chosen
multi-label evaluation metric. Therefore, when taking into account the evaluation metric, the notion of
confidence is not clear. Expected accuracy provides the
practitioner with a measurement of the estimated accuracy with respect
to the evaluation metric.

We hypothesised that expected accuracy of a multi-label prediction is
a function of the categorical distribution over the powerset of the
labels conditioned on the given observation.  In this article we
examined the correlation and robustness of seven candidate functions
for measuring the expected accuracy of each prediction using the
multi-label distribution. We found that three of the candidate
functions were indeed highly correlated to expected accuracy and were also
robust to changes in variables (such as changes in data).

When fitting a model of expected accuracy with respect to different
similarity (performance, evaluation) metrics, we found that each of the three candidate functions were
appropriate for predicting Hamming expected accuracy, but a
combination of the functions was better suited to estimating Jaccard
and Exact expected accuracy.

\bibliographystyle{plainnat}
\bibliography{acceptanceCandidatesAX}

\appendix

\section{Situational Correlation}
\label{sec:rawcor}

Relative acceptance implies that the order provided by accuracy and
acceptance should be the same; i.e., highly correlated. To measure this correlation (i.e., ordinal similarity), we use 
Kendall's $\tau$ correlation.  Results are presented in Tables \ref{tab:correlationExact},
\ref{tab:correlationHamming}, and \ref{tab:correlationJaccard}
for the different metrics. Each table presents the correlation for the given
similarity function broken down by data set and method.

\begin{table}[ht]
\centering
  \begin{tabularx}{0.7\linewidth}{lZ@{}lY@{}lY@{}lY@{}l}
    \toprule
    Method & \multicolumn{2}{c}{HP, ME} & \multicolumn{2}{c}{TG} & \multicolumn{2}{c}{SE} & \multicolumn{2}{c}{CE, CS, GI} \\
  \midrule
  \textit{Emotion} \\ \zspace{}ECC & 0.2595 &  & 0.1695 & \downthree & 0.2345 &  & 0.2523 &  \\
  \zspace{}CT & 0.2658 &  & 0.1442 & \downtwo & 0.2133 & \downtwo & 0.2362 & \downone \\
  \zspace{}Indep & 0.1272 &  & 0.0409 & \downtwo & 0.1636 &  & 0.1568 &  \\
\textit{Scene} \\ \zspace{}ECC & 0.3440 &  & 0.3365 &  & 0.3033 & \downthree & 0.3276 & \downthree \\
  \zspace{}CT & 0.3073 &  & 0.2676 & \downthree & 0.3424 & \upthree & 0.3222 & \uptwo \\
  \zspace{}Indep & 0.2690 &  & 0.2106 & \downthree & 0.3407 & \upthree & 0.2994 & \upthree \\
\textit{Yeast} \\ \zspace{}ECC &  0.3167 &  &  0.3066 &  &  0.2789 & \downthree &  0.2949 & \downthree \\
  \zspace{}CT &  0.3117 &  &  0.1978 & \downthree &  0.3126 &  &  0.3171 &  \\
  \zspace{}Indep &  0.3217 &  &  0.2448 & \downthree &  0.2548 & \downthree &  0.3383 & \uptwo \\
\textit{Stare} \\ \zspace{}ECC & 0.2651 &  & 0.2506 &  & 0.1813 & \downtwo & 0.2384 &  \\
  \zspace{}CT & 0.2452 &  & 0.2079 &  & 0.2821 &  & 0.2490 &  \\
  \zspace{}Indep & 0.0889 &  & 0.0526 & \downthree & 0.0927 &  & 0.0939 & \upone \\
\textit{Enron} \\ \zspace{}ECC &  0.3223 &  &  0.2983 & \downtwo &  0.2694 & \downtwo &  0.3242 &  \\
  \zspace{}CT &  0.3122 &  &  0.2904 & \downone &  0.2662 & \downone &  0.3049 &  \\
  \zspace{}Indep &  0.3191 &  &  0.2981 & \downtwo &  0.3483 &  &  0.3504 & \uptwo \\
\textit{Slashdot} \\ \zspace{}ECC &  0.1886 &  &  0.1018 & \downthree &  0.2618 & \upthree &  0.2472 & \upthree \\
  \zspace{}CT &  0.1030 &  &  0.0219 & \downthree &  0.2316 & \upthree &  0.1652 & \upthree \\
  \zspace{}Indep &  0.0546 &  & -0.0208 & \downthree &  0.1993 & \upthree &  0.1196 & \upthree \\
   \bottomrule
\end{tabularx}
\caption{Correlation between the chosen acceptance function and the Exact Match accuracy for each data set and each multi-label method. Arrows show a significant difference at the 0.1 (one arrow), 0.05 (two arrows) and 0.01 (three arrows) levels, when compared to HP.} 
  \label{tab:correlationExact}
\end{table}

\begin{table}[ht]
\centering
  \begin{tabularx}{0.7\linewidth}{lZ@{}lY@{}lY@{}lY@{}l}
    \toprule
    Method & \multicolumn{2}{c}{HP, ME} & \multicolumn{2}{c}{TG} & \multicolumn{2}{c}{SE} & \multicolumn{2}{c}{CE, CS, GI} \\
  \midrule
  \textit{Emotion} \\ \zspace{}ECC & 0.2527 &  & 0.1549 & \downthree & 0.2280 &  & 0.2434 &  \\
  \zspace{}CT & 0.2245 &  & 0.0506 & \downthree & 0.2013 &  & 0.2212 &  \\
  \zspace{}Indep & 0.1223 &  & 0.0216 & \downthree & 0.1593 & \upone & 0.1503 &  \\
  \textit{Scene} \\ \zspace{}ECC & 0.346 &  & 0.331 & \downone & 0.302 & \downthree & 0.330 & \downthree \\
  \zspace{}CT & 0.302 &  & 0.260 & \downthree & 0.328 & \uptwo & 0.315 & \uptwo \\
  \zspace{}Indep & 0.264 &  & 0.205 & \downthree & 0.320 & \upthree & 0.290 & \upthree \\
  \textit{Yeast} \\ \zspace{}ECC &  0.2948 &  &  0.2610 & \downtwo &  0.2733 & \downone &  0.2859 &  \\
  \zspace{}CT &  0.2492 &  &  0.1176 & \downthree &  0.2634 &  &  0.2668 & \uptwo \\
  \zspace{}Indep &  0.2399 &  &  0.1528 & \downthree &  0.1534 & \downthree &  0.2719 & \upthree \\
  \textit{Stare} \\ \zspace{}ECC & 0.2818 &  & 0.2497 &  & 0.2301 &  & 0.2738 &  \\
  \zspace{}CT & 0.2040 &  & 0.1537 & \downtwo & 0.2469 &  & 0.2195 &  \\
  \zspace{}Indep & 0.1568 &  & 0.0958 & \downtwo & 0.1738 &  & 0.1662 &  \\
  \textit{Enron} \\ \zspace{}ECC &  0.3204 &  &  0.2644 & \downthree &  0.3200 &  &  0.3568 & \uptwo \\
  \zspace{}CT &  0.3023 &  &  0.2513 & \downthree &  0.2988 &  &  0.3261 &  \\
  \zspace{}Indep &  0.3188 &  &  0.2640 & \downthree &  0.3951 & \upthree &  0.3842 & \upthree \\
  \textit{Slashdot} \\ \zspace{}ECC &  0.2003 &  &  0.1236 & \downthree &  0.2537 & \upthree &  0.2493 & \upthree \\
  \zspace{}CT &  0.1220 &  &  0.0493 & \downthree &  0.2256 & \upthree &  0.1762 & \upthree \\
  \zspace{}Indep &  0.0778 &  &  0.0107 & \downthree &  0.1927 & \upthree &  0.1336 & \upthree \\
   \bottomrule
\end{tabularx}
\caption{Correlation between the chosen acceptance function and Hamming accuracy for each data set and each multi-label method. Arrows show a significant difference at the 0.1 (one arrow), 0.05 (two arrows) and 0.01 (three arrows) levels, when compared to HP.} 
  \label{tab:correlationHamming}
\end{table}

\begin{table}[ht]
\centering
  \begin{tabularx}{0.7\linewidth}{lZ@{}lY@{}lY@{}lY@{}l}
    \toprule
    Method & \multicolumn{2}{c}{HP, ME} & \multicolumn{2}{c}{TG} & \multicolumn{2}{c}{SE} & \multicolumn{2}{c}{CE, CS, GI} \\
  \midrule
  \textit{Emotion} \\ \zspace{}ECC & 0.3046 &  & 0.1692 & \downthree & 0.2700 & \downone & 0.2945 &  \\
  \zspace{}CT & 0.2989 &  & 0.0694 & \downthree & 0.2491 & \downtwo & 0.2857 &  \\
  \zspace{}Indep & 0.1927 &  & 0.0709 & \downthree & 0.2115 &  & 0.2138 &  \\
  \textit{Scene} \\ \zspace{}ECC & 0.348 &  & 0.335 & \downone & 0.311 & \downthree & 0.333 & \downthree \\
  \zspace{}CT & 0.310 &  & 0.259 & \downthree & 0.356 & \upthree & 0.329 & \upthree \\
  \zspace{}Indep & 0.263 &  & 0.194 & \downthree & 0.349 & \upthree & 0.298 & \upthree \\
  \textit{Yeast} \\ \zspace{}ECC &  0.2865 &  &  0.2521 & \downthree &  0.2634 & \downone &  0.2761 &  \\
  \zspace{}CT &  0.2762 &  &  0.1295 & \downthree &  0.2914 &  &  0.2976 & \uptwo \\
  \zspace{}Indep &  0.2586 &  &  0.1449 & \downthree &  0.1306 & \downthree &  0.3051 & \upthree \\
  \textit{Stare} \\ \zspace{}ECC & 0.2941 &  & 0.2828 &  & 0.1992 & \downtwo & 0.2633 &  \\
  \zspace{}CT & 0.2139 &  & 0.1778 &  & 0.2993 & \upthree & 0.2413 &  \\
  \zspace{}Indep & 0.1377 &  & 0.0879 &  & 0.2073 & \uptwo & 0.1620 &  \\
  \textit{Enron} \\ \zspace{}ECC & -0.1284 &  & -0.1160 &  &  0.0648 & \upthree & -0.0509 & \upthree \\
  \zspace{}CT & -0.1238 &  & -0.1125 &  &  0.1254 & \upthree & -0.0333 & \upthree \\
  \zspace{}Indep & -0.1509 &  & -0.1231 &  & -0.0628 & \upthree & -0.1115 &  \\
  \textit{Slashdot} \\ \zspace{}ECC &  0.1802 &  &  0.0892 & \downthree &  0.2671 & \upthree &  0.2442 & \upthree \\
  \zspace{}CT &  0.0849 &  &  0.0010 & \downthree &  0.2339 & \upthree &  0.1541 & \upthree \\
  \zspace{}Indep &  0.0329 &  & -0.0445 & \downthree &  0.1954 & \upthree &  0.1040 & \upthree \\
   \bottomrule
\end{tabularx}
\caption{Correlation between the chosen acceptance function and Jaccard accuracy for each data set and each multi-label method. Arrows show a significant difference at the 0.1 (one arrow), 0.05 (two arrows) and 0.01 (three arrows) levels, when compared to HP.} 
  \label{tab:correlationJaccard}
\end{table}

The results show that TG is either equivalent to, or worse that
HP in all cases. Also we note that Shannon Entropy
and the Collision Entropy equivalents are superior under \dataset{Slashdot}. 
For all other combinations, the results are mixed; there are
cases when HP has the highest correlation and cases where the
other candidate acceptance function have higher correlation.  

We conclude that candidate acceptance
functions other than HP are more appropriate under given
conditions. 

\begin{figure*}
  \hfill
\includegraphics[scale=0.44]{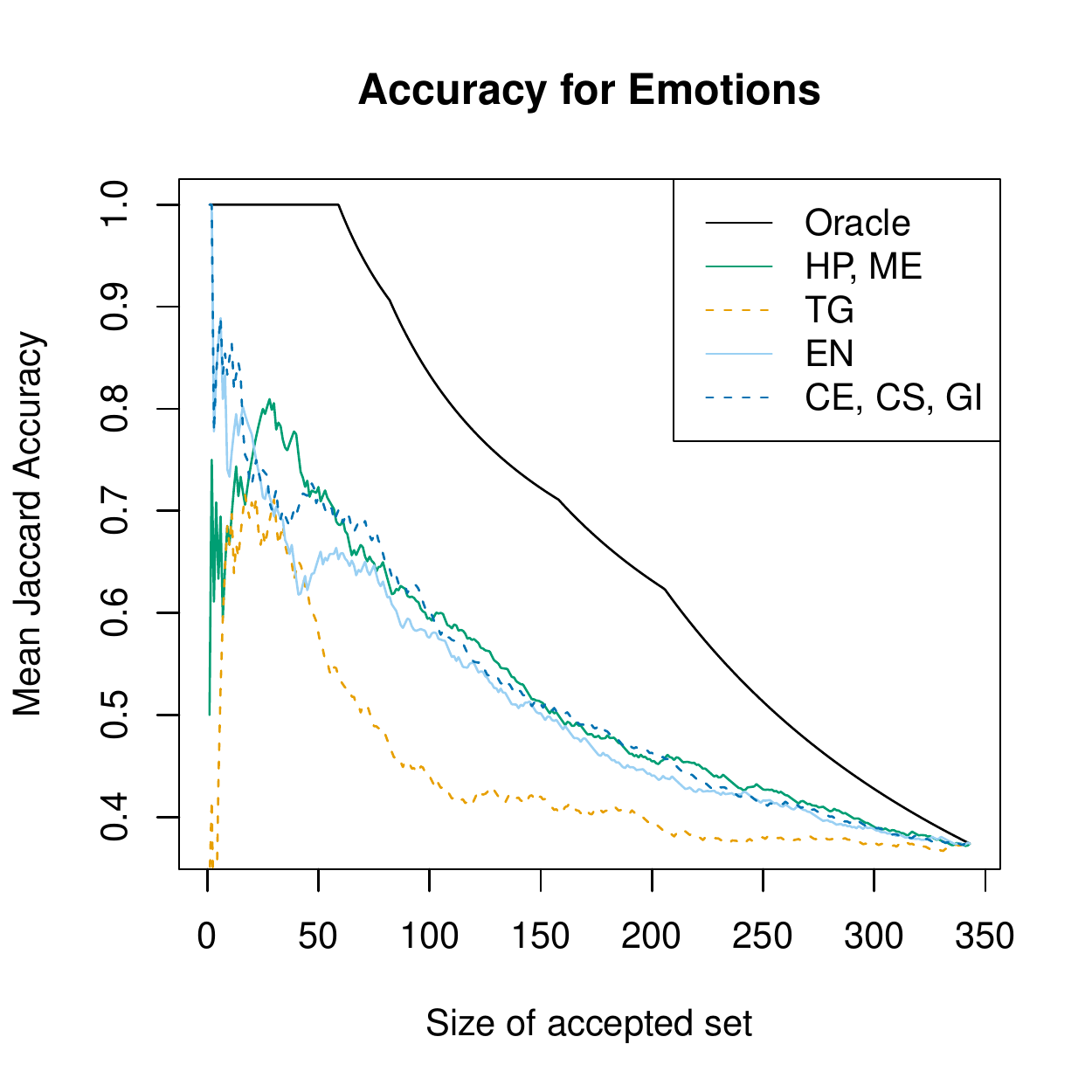} \hfill
\includegraphics[scale=0.44]{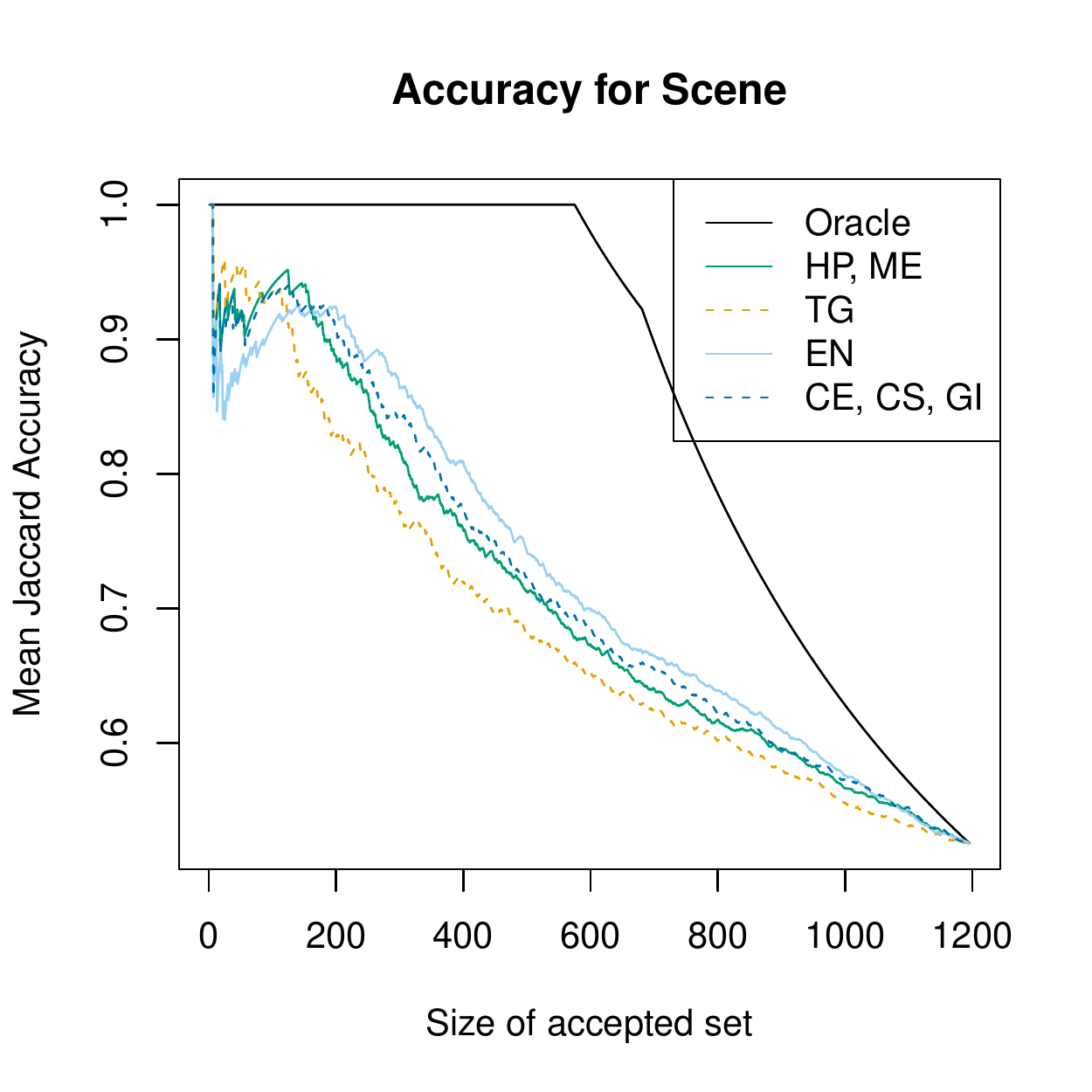} \hfill ~ \\

~\hfill
\includegraphics[scale=0.44]{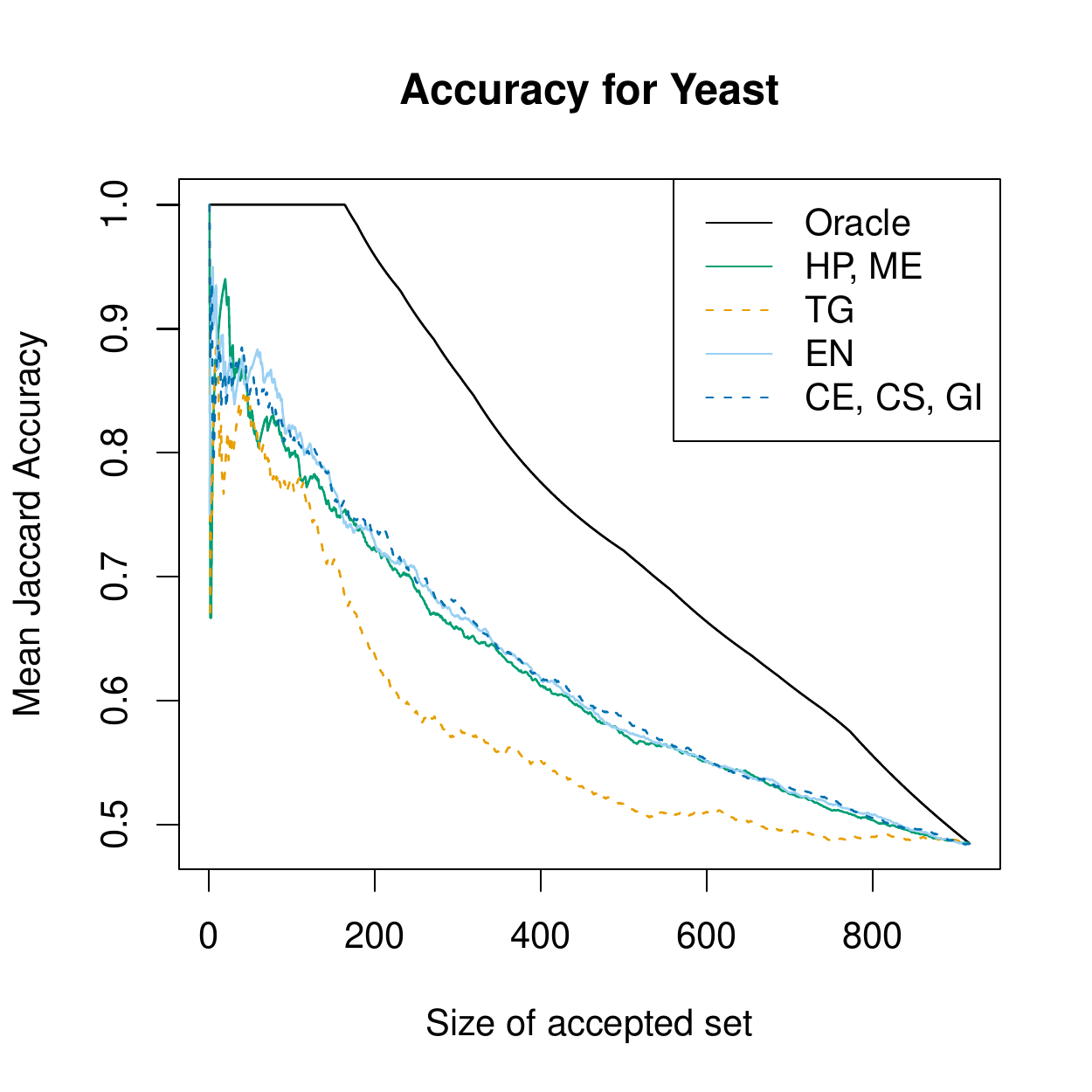} \hfill
\includegraphics[scale=0.44]{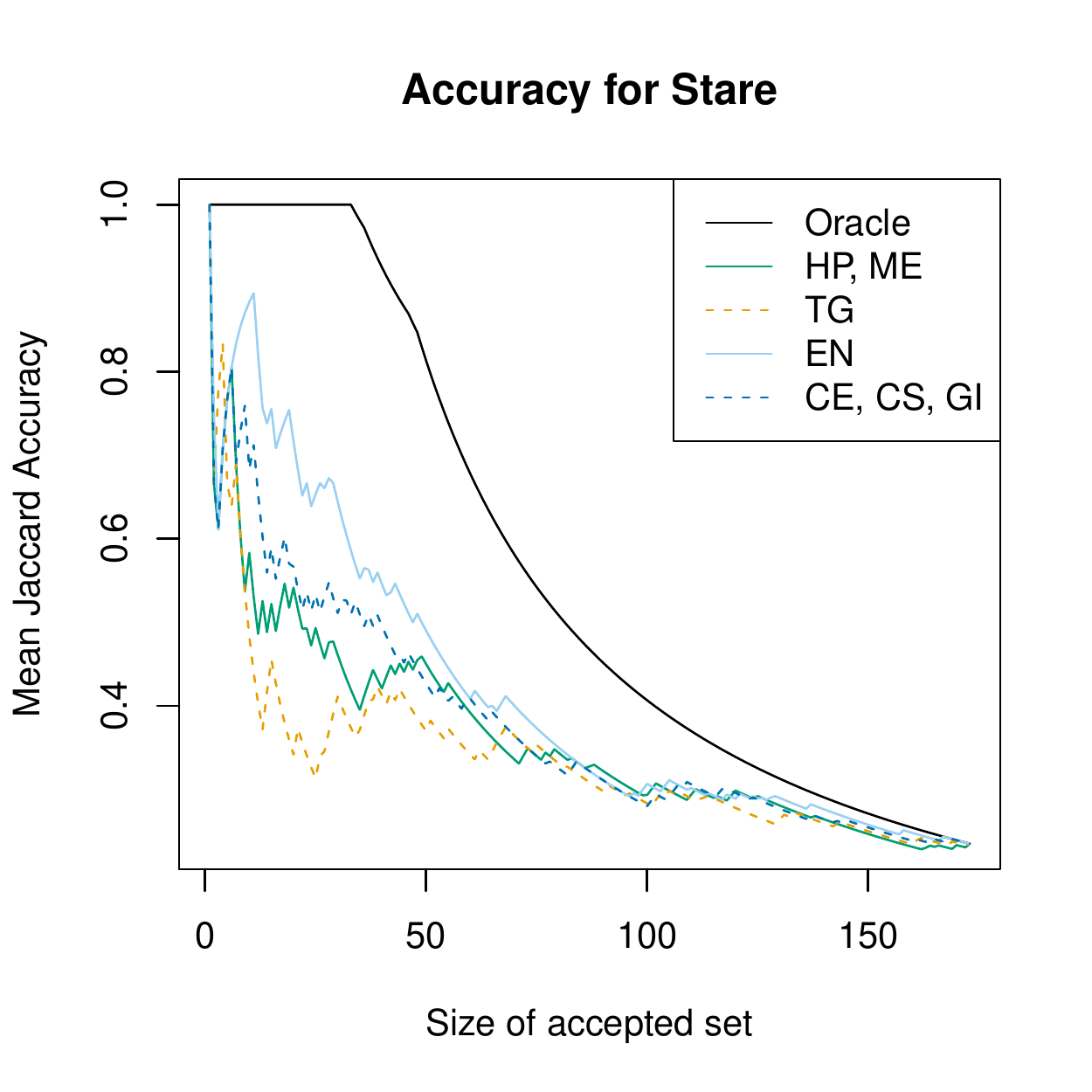} \hfill ~ \\

~\hfill
\includegraphics[scale=0.44]{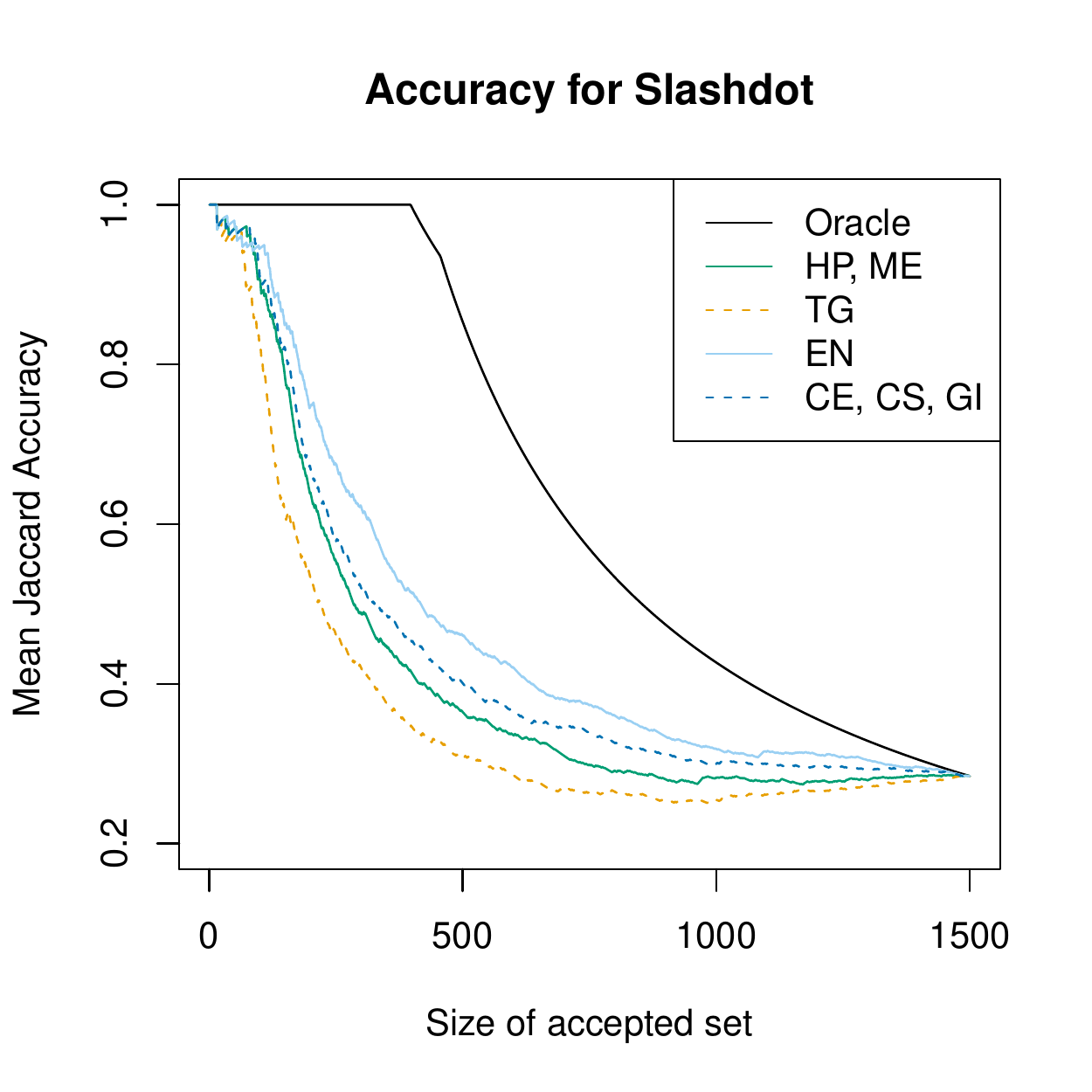} \hfill
\includegraphics[scale=0.44]{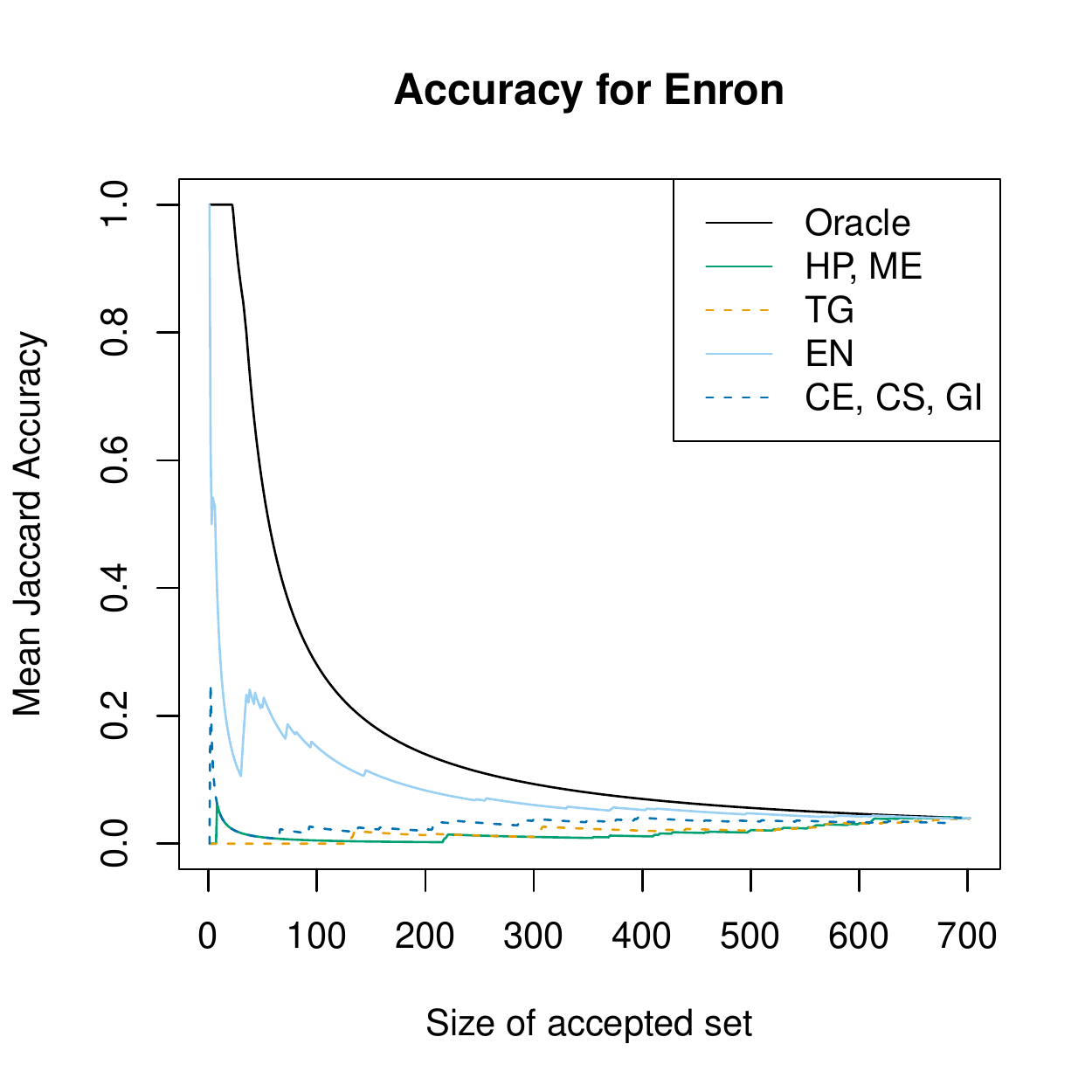} \hfill ~
\caption{Mean Jaccard accuracy when choosing only the predictions with
  greatest acceptance score when using a Classifier Trellis over the
  six data sets. The $x$-axis shows the number of predictions accepted
  (e.g. 1 implies that only the prediction with the greatest
  acceptance score was included in the mean, and the right most value
  is when including all predictions in the mean accuracy).}
\label{fig:initialplots}
\end{figure*}

This is reinforced by Figure \ref{fig:initialplots}, which provides a graphical representation of
the CT method results from Table \ref{tab:correlationJaccard}.

\end{document}